\documentclass[10pt,twocolumn,letterpaper]{article}

\usepackage{iccv}
\usepackage{times}
\usepackage{epsfig}
\usepackage{graphicx}
\usepackage{amsmath}
\usepackage{amssymb}
\usepackage{color}
\usepackage{xcolor}
\usepackage{authblk}
\usepackage{multicol}
\usepackage[accsupp]{axessibility} 
\makeatletter
\renewcommand\@author{
    \AB@authlist\\
    \begin{multicols}{2}
      \begin{quotation}
      {\setlength{\parindent}{0cm}
      \noindent
        \AB@affillist
        }
      \end{quotation}
    \end{multicols}
    }

\usepackage[pagebackref=true,breaklinks=true,letterpaper=true,colorlinks,bookmarks=false]{hyperref}

\iccvfinalcopy 

\def\iccvPaperID{7544} 

\ificcvfinal\fi

\newcommand{\myparagraph}[1]{\vspace{3pt}\noindent{\bf{#1}}}

\begin{document}

\title{PDiscoNet: Semantically consistent part discovery for fine-grained recognition}

\author[ ]{Robert van der Klis$^1$ \hspace{10pt} Stephan Alaniz$^2$\hspace{10pt}  Massimiliano Mancini$^{3}$\hspace{10pt}  Cassio F. Dantas$^{4,6}$\hspace{10pt}  \\ Dino Ienco$^{4,6}$\hspace{10pt}  Zeynep Akata$^{2}$\hspace{10pt}  Diego Marcos$^{5,6}$\\
$^1$WUR\hspace{6pt}  $^2$University of T\"ubingen\hspace{6pt}  $^3$University of Trento\hspace{8pt}  $^4$INRAE/UMR TETIS\hspace{6pt}  $^5$Inria\hspace{6pt}  $^6$University of Montpellier}

\maketitle
\ificcvfinal\thispagestyle{empty}\fi

\begin{abstract}
Fine-grained classification often requires recognizing specific object parts, such as beak shape and wing patterns for birds. Encouraging a fine-grained classification model to first detect such parts and then using them to infer the class could help us gauge whether the model is indeed looking at the right details better than with interpretability methods that provide a single attribution map.
We propose PDiscoNet to discover object parts by using only image-level class labels along with priors encouraging the parts to be: discriminative, compact, distinct from each other, equivariant to rigid transforms, and active in at least some of the images. In addition to using the appropriate losses to encode these priors, we propose to use part-dropout, where full part feature vectors are dropped at once to prevent a single part from dominating in the classification, and part feature vector modulation, which makes the information coming from each part distinct from the perspective of the classifier.
Our results on CUB, CelebA, and PartImageNet show that the proposed method provides substantially better part discovery performance than previous methods while not requiring any additional hyper-parameter tuning and without penalizing the classification performance. The code is available at \url{https://github.com/robertdvdk/part_detection}
\end{abstract}

\section{Introduction}

Commonly used approaches to inspect a deep learning model's inner workings yield a saliency map that indicates which regions contributed the most to the output~\cite{bach2015pixel,selvaraju2017grad}.
If the model seems to focus on image regions that are known to be irrelevant, (e.g. the background or the wrong object), it becomes clear that the model has picked up on spurious correlations and cannot be trusted. This observation could then be used to improve future iterations of the model, for instance, by eliminating or compensating for the detected spurious correlations.
However, this type of approach offers little information when the model provides an incorrect answer but the saliency map suggests that it is attending to the correct image regions.

\begin{figure}[t]
    \centering
    \includegraphics[width=\columnwidth]{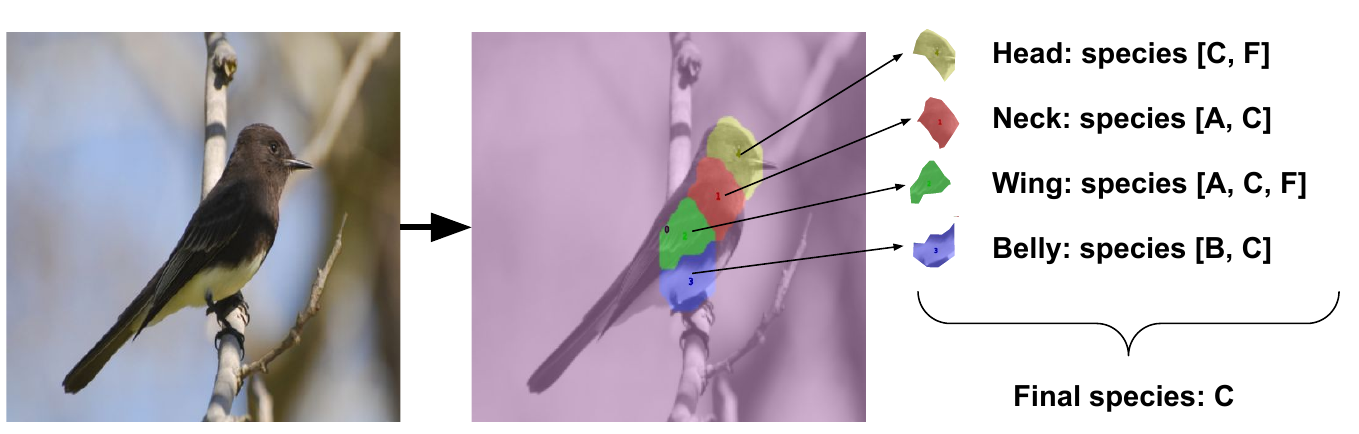}
    \caption{Our PDiscoNet extracts semantically consistent parts, without any part annotations, and reasons on these parts before combining the results into a final fine-grained classification output.}
    \label{fig:splash}
\end{figure}

Other approaches aim at modifying the model architecture itself in order to ensure that the provided explanation actually reflects the decision process of the model~\cite{chen2019looks,bohle2021convolutional}.
In particular, the saliency map explanation can be enriched by dividing it in multiple semantically interpretable parts, mimicking the traditional approaches of tackling fine-grained visual categorization (FGVC), in which image-level part annotations were leveraged~\cite{krause2015fine} in order to help the model differentiate between similar classes by helping it focus on the relevant parts. In this manner, we have more information to judge the adequacy of the model's reasoning: even if the correct object is highlighted, we will be suspicious of the result if the part map that the model generally associated to the head of a bird seems to highlight the feet in one particular image.
We thus posit that a model that classifies images based on just a few discriminative regions that are semantically consistent across images would be more interpretable than one which highlights the whole object, as one can immediately visualise the parts of the image that have been attended to and interpret their semantics across images. By inspecting a few images and their corresponding detected parts, we can easily assign semantic meaning to each part (e.g., bird beak, vehicle wheel) and judge whether the correct parts are being detected in a new image. 

Even if the model correctly assigns high saliency to the object of interest, we will know to mistrust the result in case the discovered part semantics are not respected.
This way of interpreting the models has an additional advantage over post-hoc methods in that we can be more certain that the model only uses information from the indicated regions. Such models have also been shown to be more robust; irrelevant parts are filtered out by only looking at the discovered discriminative regions, which can have a positive impact on generalization capability and thus robustness to occlusion~\cite{zhang2018deepvoting} and adversarial attacks~\cite{sitawarin2022part}.

Discovering  meaningful and discriminative parts using only image-level class labels requires the use of additional priors that encode our expectations on the characteristics of these parts along with a model architecture that allows for these priors to be implemented.
We design a model, based on a Convolutional Neural Network (CNN) backbone, which discovers discriminative parts of objects by being forced to use the discovered parts as a bottleneck for fine-grained classification. The fine-grained setting ensures a high level of similarity between classes, enabling the possibility of discovering semantic parts that are shared by multiple classes.
In our part bottleneck, class logits are independently extracted from each of the discovered parts before being combined for the final classification, along with a dropout layer that affects whole parts at a time, ensures that all discovered parts are relevant to classification.


\section{Related work}

\myparagraph{Fine-grained recognition}
FGVC is a classification setting in which objects of multiple sub-classes of the same super-class are present, thus constituting a challenging task where subtle intra-class and large inter-class variation need to be simultaneously addressed~\cite{WeiSAWPTYB22}.
Solving fine-grained tasks usually requires one to closely inspect the object for the telltale differences between closely related classes. Traditional methods exploit shared keypoints~\cite{krause2014learning,krause2015fine}, parts~\cite{HuangXTZ16,YinWZ20}, attributes~\cite{LiuLQWT16,LiuWWDL17}, or a pre-segmentation of the object of interest~\cite{angelova2013efficient} in order to effectively discriminate between similar sub-classes, although deep learning approaches using large quantities of data have since also proved effective~\cite{krause2016unreasonable}.
Our PDiscoNet belongs to a family of approaches that facilitate injecting some of the structure provided by part-based~\cite{GeLY19,huang_interpretable_2020} or attribute-based~\cite{duan2012discovering} reasoning via weakly-supervised learning without part or attribute annotations.

\myparagraph{Interpretability via attribution maps} In saliency-based attribution methods, the goal is to highlight important regions of the image that are used by the network to form its decision. Examples include perturbation-based~\cite{LundbergL17,PetsiukDS18}, activation-based~\cite{KimWGCWVS18,ZhouKLOT16}, and gradient-based~\cite{selvaraju2017grad,SundararajanTY17} explanations methods. Despite their popularity, these model-agnostic methods often cannot guarantee that their explanations are faithful to the model~\cite{AdebayoGMGHK18}. In contrast, inherently interpretable models aim to directly expose the decision process of the network~\cite{BohleFS22,chen2019looks}. In this work, we focus on incorporating interpretable components into the network architecture to reveal the learned structure transparently. Attention rollout~\cite{AbnarZ20,CheferGW21} is a popular way to understand whether attention modules can provide such explanations~\cite{JainW19,WiegreffeP19}. However, deep transformer architectures model complex functions such that reliable interpretation is often limited to inspecting single self-attention layers~\cite{CaronTMJMBJ21,amir_deep_2022_dinovit}. Based on this observation we employ a shallow attention structure into our network that allows to directly explain the attention maps with the correspondence to object parts.

\myparagraph{Unsupervised part discovery}
Some previous works discover parts by using image reconstruction~\cite{zhang2018unsupervised,choudhury_unsupervised_2022},
where a landmark bottleneck is used to discover object parts.
However, the model having no learning signal indicating which parts of the image may represent an object of interest limits the applicability of these approaches to cases where the objects of interest are either dominating the image and depicted in similar poses~\cite{zhang2018unsupervised} or are endowed with foreground segmentation masks that can be used as an additional training signal~\cite{choudhury_unsupervised_2022}.
On datasets where most parts are common to all images, pre-trained Vision Transformer~\cite{amir_deep_2022_dinovit} is typically able to find the parts of the most relevant object in a semantically consistent manner. However, it breaks down when the assumption that salient parts occur in almost all images does not hold, since parts tend to become polysemous in such a setting.
Unlike these approaches, PDiscoNet is able to leverage the class labels, requiring no additional annotations in fine-grained classification datasets, to learn parts that are specific to similar classes, making them more semantically consistent and suitable for interpretation.

\begin{figure*}
  \includegraphics[width=\textwidth]{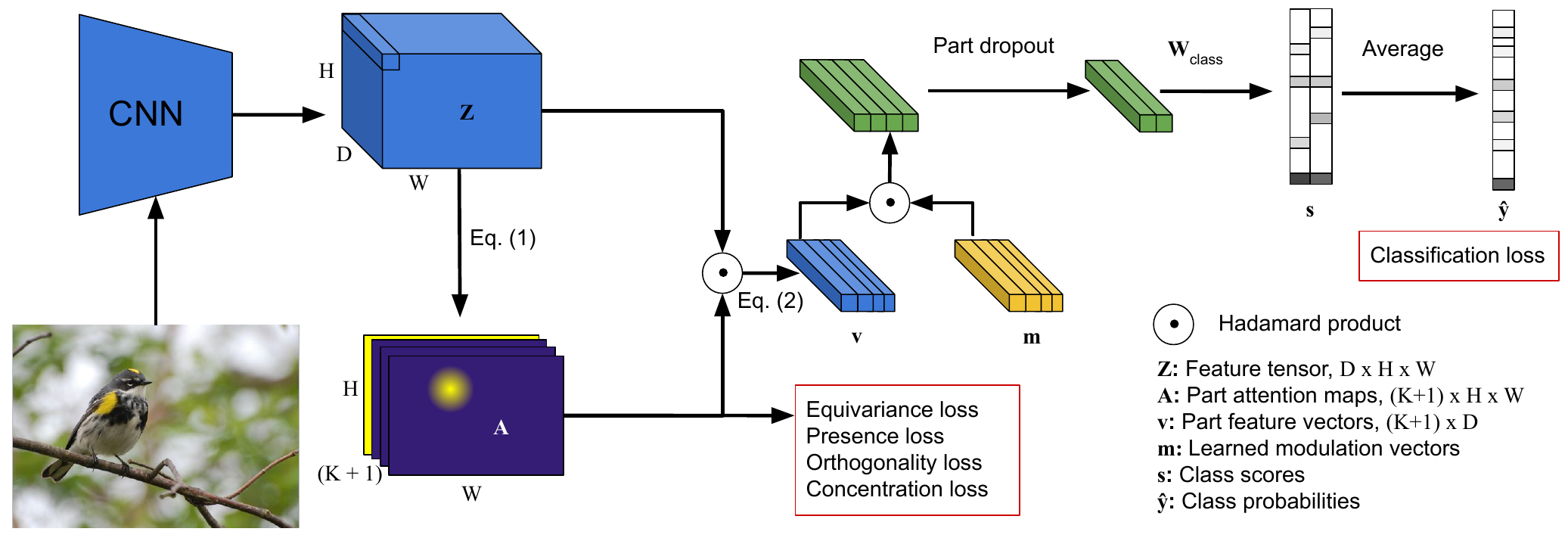}
  \caption{Diagram of the proposed method. The part discovery process is driven by the fine-grained classification loss and the losses that applied on the part attention maps (red boxes).}
\end{figure*}

\myparagraph{Weakly-supervised part discovery via FGVC}
MA-CNN~\cite{zheng2017learning} and ProtoPNet~\cite{chen2019looks} propose to directly enforce that the CNN activation maps develop a part-like behaviour, 
showing that an architecture with enhanced interpretability does not result in a loss of performance. However, their focus is more on downstream fine-grained classification than on evaluating the discovered parts.
SCOPS~\cite{hung_scops_2019}, a model for part co-segmentation, puts more emphasis on the quality of the discovered parts by adding several losses on the part maps that encourage them to be compact and distinct, the latter 
via decorrelation of the learned part prototypes. It also encourages part maps to be equivariant under geometric transforms of the image. Taken together, these incentives ensure that the discovered parts are semantically consistent across images.
This method assumes that all the parts should be active in every image of the dataset. Huang and Li\cite{huang_interpretable_2020} aim to solve this issue by encouraging the presence of each part across a batch of images to follow a beta distribution with manually defined parameters. Depending on the chosen parameters, this encourages a pre-defined proportion of images in a batch to display the part, while it is discouraged in the rest of the images in the batch.


\section{PDiscoNet Method}
We design an approach to discover $K$ \textit{discriminative parts} that are relevant to a fine-grained classification task, based solely on the image-level class labels. Let $\mathbf{X} \in \mathbb{R}^{3\times A\times B}$ denote an image in the dataset, and let $y \in \{1, 2, ..., C\}$ be its corresponding label.
Using a CNN base model $f_\theta$ we obtain a feature tensor $\mathbf{Z}=f_\theta(\mathbf{X})$ with $\mathbf{Z} \in \mathbb{R}^{D\times H\times W}$.
Following~\cite{hung_scops_2019} and~\cite{huang_interpretable_2020}, from this tensor we compute $K+1$ ($K$ parts plus one background element) attention maps $\mathbf{A}^k = [0, 1]^{H\times W},\ k \in \{1,\dots,K+1\}$ by applying a negative squared Euclidean distance function between feature vectors $\mathbf{z}_{ij}$ (with $\mathbf{z}_{ij} \in \mathbb{R}^D$, $i \in \{1, ..., H\}$, $j \in \{1, ..., W\}$) and $K$ part prototypes $\mathbf{p}^k\in\mathbb{R}^D$ in a $1\times 1$ convolutional manner, followed by a softmax across the $K+1$ channels:
\begin{equation}
    a_{ij}^k = \frac{\exp(-\|\mathbf{z}_{ij}-\mathbf{p}^k\|^2)}{\sum_k \exp(-\|\mathbf{z}_{ij}-\mathbf{p}^k\|^2)},
\end{equation}
Each attention map is then used to compute its corresponding part vector $\textbf{v}^k \in \mathbb{R}^D$ by using the attention values to calculate a weighted average over the feature vectors in $\mathbf{Z}$:
\begin{equation}
    \mathbf{v}^k=\frac{\sum_i\sum_j \mathbf{z}_{ij} a_{ij}^k}{HW}
\end{equation}

Each of these part vectors could then be used to obtain a vector of class scores $\mathbf{s}^k\in\mathbb{R}^C$ calculated as 
$$ \mathbf{s}^k = \mathbf{W}_{class}\mathbf{v}^k$$
by applying the same linear classifier $\mathbf{W}_{class}\in\mathbb{R}^{C\times D}$ to all part feature vectors, but we use the modification in Eq.~(\ref{eq:mod}). The scores are then averaged into a single score vector $\mathbf{s} = \frac{1}{K}\sum_k \mathbf{s}^k$ on which a softmax is applied to obtain the final classification probabilities $\mathbf{\hat{y}}$.


\myparagraph{Part vector modulation}
In the above formulation, all parts share the same classifier weights $W_\text{class}$. This poses the problem that, from the perspective of the classifier, all parts are equivalent, meaning that the classifier could be encouraging all parts of the same object to result in the similar feature representation. The classifier can also profit from part misdetection, since a wrongly detected part would still provide a useful feature vector.
Although it would, in principle, be possible to learn part-specific classifiers, this would not scale well to fine-grained classification scenarios where the classification head already contains the majority of learnable weights.
As an alternative, we propose to keep a modulation vector $\mathbf{m}_k\in\mathbb{R}^D$ per landmark that multiplies element-wise each part vector before classification:
\begin{equation}
    \textbf{s}^k = W_{\text{class}}\cdot(\textbf{m}_k\odot\textbf{v}_k).
    \label{eq:mod}
\end{equation}

\myparagraph{Part dropout}
We would like the learned parts to be as discriminative as possible. In order to prevent the most discriminative parts (such as the head in birds) to discourage other parts from becoming discriminative by rendering them unnecessary, we propose to randomly drop out a proportion of all parts during training. This encourages the model to find a variety of discriminative parts.

\myparagraph{Loss functions}
The main learning signal for our model comes from fine-grained classification, for which we use cross-entropy on the output classification probabilities $\mathcal{L}_\text{class}(\mathbf{y},\hat{\mathbf{y}})$. Although this signal itself would suffice for the model to perform well in the classification task, it does not guarantee that the learned attention maps will be interpretable as parts. There are several desirable properties we wish to enforce in the learned parts. First, parts must be discriminative. This is taken care of by the classification as described previously. However, we also wish parts to be:\newline
\textbf{Compact} ($\mathcal{L}_\text{conc}$): We would like each detected part to consist of a compact and contiguous image region.\newline
\textbf{Distinct} ($\mathcal{L}_\text{orth}$): We want to avoid overlap between parts. This is encouraged by decorrelating part feature vectors.\newline
\textbf{Consistent} ($\mathcal{L}_\text{equiv}$): The same parts should be detected under translation, rotation or scaling of the image. This can be enforced via a loss that encourages the equivariance of the attention maps to random rigid transforms.\newline
\textbf{Present in the dataset} ($\mathcal{L}_\text{pres}$): All parts should be present in some of the images of the dataset. For this, we penalize the absence of a part across a whole batch during training.

To enforce these priors, we use as many loss functions.
%
Our concentration loss over the attention maps $\mathbf{A}^k$:
\begin{align}
    \mathcal{L}_\text{conc} = \frac{\sum_{k=1}^K\sigma_v^2(\mathbf{A}^k) + \sigma_h^2(\mathbf{A}^k)}{K},
\end{align}
where $\sigma_v$ and $\sigma_h$ represent the vertical and horizontal spatial variance respectively.

We calculate an orthogonality loss over the part vectors by applying the cosine distance between all pairs:
\begin{equation}
    \mathcal{L}_\text{orth}  = \sum_k\sum_{l\neq k} \frac{\mathbf{v}^{k} \cdot \mathbf{v}^l} {\|\mathbf{v}^{k}\| \cdot \| \mathbf{v}^l\|}.
\end{equation}
Our equivariance loss creates a transformed image by applying a random rigid transformation $T$ to the input image. We then pass both the original and the transformed image through the model and invert the transformation on the attention maps from the transformed ones. If $A^{k}(\mathbf{X})$ is a function that returns the $k^{th}$ attention map for image $\mathbf{X}$, the equivariance loss is computed using the cosine distance between the attention maps from the original image and the transformed image:

\begin{equation}
      \mathcal{L}_\text{equiv}  = 1- \frac{1}{K}\sum_k \frac{ \big\| A^{k}(\mathbf{X}) \odot T^{-1}(A^{k}(T(\mathbf{X}))) \big\|} {\|A^{k}(\mathbf{X})\| \cdot \| A^{k}(T(\mathbf{X}))\|}.
\end{equation}

Lastly, a presence loss encourages each part to be present at least once per batch. Given a batch $\{\mathbf{X}_1,\dots,\mathbf{X}_B\}$:
\begin{equation}
    \mathcal{L}_\text{pres} = 1 - \frac{1}{K}\sum_k\max_{b,i,j}\text{avgpool}(a_{ij}^k(\mathbf{X}_b)),
\end{equation}
where $\text{avgpool}()$ is a 2D average pooling with a small kernel size and a stride of 1. This operator is applied to prevent encouraging single pixel attention maps.
A weighted combination of these losses is used as the final loss.

\section{Experiments}

We compare our method, for different values of $K$, against the results obtained by the most closely related methods in the recent literature~\cite{huang_interpretable_2020}.
We also compare our method to a few other methods, among which the most recent method on part discovery~\cite{amir_deep_2022_dinovit}, which is not aimed at fine-grained classification but showcases high quality part discovery by using self-supervised pretraining with a visual transformer architecture.

\myparagraph{Datasets}
Our aim is to perform part discovery with the only assumption being that we have image-level class labels where parts are shared by some of the classes, which is typically the case in FGVC tasks. In order to investigate this, we have chosen three datasets with a varying proportions of shared parts across images: a face image dataset where the vast majority of images display all relevant parts (\emph{i.e.} facial landmarks), a bird species recognition dataset, where the assumption of the presence of all parts is limited due to the effects of pose and occlusion, and a more challenging dataset in which several fine-grained class categories (\emph{e.g.} birds and cars) are mixed together, resulting in specific parts only being shared by a small subset of the images in the dataset.
To assess the quality of the discovered parts, we have selected datasets for which semantic part annotations are available.

\begin{table*}[t]
\centering
\begin{tabular}{l||ccc|c||ccc||cc|c}
 & \multicolumn{4}{c||}{\textbf{CUB}} & \multicolumn{3}{c||}{\textbf{CelebA}} & \multicolumn{3}{c}{\textbf{PartImageNet}}\\
             & Kp.$\downarrow$ & NMI$\uparrow$  & ARI$\uparrow$   & Class.$\uparrow$ 
             & Kp. reg. $\downarrow$ & NMI$\uparrow$    & ARI$\uparrow$
             & NMI$\uparrow$    & ARI$\uparrow$  & Class. $\uparrow$  \\ \hline\hline
Choudhury \cite{choudhury_unsupervised_2022}$^*$    & 9.20                & 43.50 & 19.60 & -
& - & - & - & - & - & - \\ \hline\hline
SCOPS \cite{hung_scops_2019}$^{**}$       & 12.60               & 24.40 & 7.10  & -   
      & 15.00    &   -    &  -  & - & - & -  \\
DFF~\cite{collins2018deep}$^{**}$          & -                   & 25.90 & 12.40 & -   
     & 31.30    &  -     &  -  & - & - & -  \\ \hline
Dino \cite{amir_deep_2022_dinovit}$^{**}$ (K=4)    & -   & 31.18 & 11.21 & -
& 11.36    &  1.38     &   0.01
& 19.17  &   7.59   &   -   \\
Dino \cite{amir_deep_2022_dinovit}$^{**}$ (K=8)    & -   & 47.21 & 19.76 & -
& 10.74   &  1.12     &   0.01
&   31.46  &   14.16  &   -  \\
Dino \cite{amir_deep_2022_dinovit}$^{**}$  (K=16)   & -   & 50.57 & 26.14 & -
& -    &  3.29     &   0.06
&   37.81  &    \textbf{16.50}   &  -  \\
\hline\hline

Huang \cite{huang_interpretable_2020} (K=4)  & 11.51   & 29.74 & 14.04 & 87.30
  & 8.75     & 56.69 & 34.74
     & 5.88 & 1.53 &  74.22 \\
Huang \cite{huang_interpretable_2020} (K=8)  & 11.60   & 35.72 & 15.90 & 86.05
 & 7.96     & 54.80 & 34.74
    & 7.56 & 1.25 &  73.56 \\
Huang \cite{huang_interpretable_2020} (K=16) & 12.60  & 43.92 & 21.10 &  85.93
 & \textbf{7.62}     & 62.22 & 41.01
   &  10.19 & 1.05 &   73.20 \\ \hline
PDiscoNet (K=4)   & 9.12                & 37.82 & 15.26 & 86.17
   & 11.11    & 75.97 & 69.53
      & 27.13 &  8.76 &  88.58\\
PDiscoNet (K=8)   & 8.52                & 50.08 & 26.96 & 86.72
  & 9.82     & 62.61 & 51.89
    & 32.41 & 10.69 &  \textbf{89.00} \\
PDiscoNet (K=16)  & \textbf{7.67}                & \textbf{56.87} & \textbf{38.05} & \textbf{87.49}
  & 9.46     & \textbf{77.43} & \textbf{70.48}
   & \textbf{41.49} & 14.17 &  86.06
\end{tabular}
\caption{Part discovery results on CUB, CelebA and PartImageNet. * Methods use foreground masks in training. ** Methods do not use class supervision. In the case of PartImageNet the number of parts are $K=[8, 25, 50]$ instead of $K=[4, 8, 16]$ used in CUB and CelebA. A ResNet101 baseline trained in the same setting as our model results in accuracies of 85.35\% on CUB and 90.81\% on PartImageNet.}
\label{tab:quant}
\end{table*}

CUB~\cite{WahCUB_200_2011} contains 11,788 images of 200 bird species that include manual part annotations of 15 body parts. The images are split approximately in half for training and half for evaluating. During the development phase of this work we used a 90\%-10\% split of the training set of CUB in order to find a good set of hyperparameters for our model and used the same across all experiments on all datasets.

CelebA~\cite{liu2015celeba} is a dataset of face images of 10,177 celebrities. We follow earlier approaches~\cite{hung_scops_2019,huang_interpretable_2020} and use the unaligned training set of 45,609 images to train our models and use the 283 MAFL test images to evaluate the part detection, and the 5,379 images of the MAFL training set were used for training the keypoint regressor. We use identity classification as the downstream task.

PartImageNet \cite{he_partimagenet_2022} consist of 158 classes split among a diverse set of categories (e.g., 10 species of fish, 14 of birds, 15 of snakes, 23 types of car). We train all models on 14,876 images of the train set, which is limited to 109 classes, and test on 1,664 images.


\myparagraph{Evaluation metrics}
The part annotations in CUB and CelebA are in the form of points, meant to represent part centroids in CUB and facial landmarks in CelebA. We first evaluate the quality of the part discovery methods by performing part location regression based on the centroids of the discovered parts.
However, as noted by~\cite{choudhury_unsupervised_2022}, keypoint regression may not be a good indicator of overall part quality. We therefore employ also the Normalized Mutual information (NMI) and the Adjusted Rand Index (ARI), metrics commonly used for evaluating clustering quality.
In PartImageNet the part annotations are in the form of semantic segmentation masks, from which we extract the centroids to compute NMI and ARI.
Note that NMI and ARI are computed on the annotation/prediction correspondences across the whole datasets, meaning that they capture part semantic consistency (i.e. a perfect score can only be obtained if the same discovered part matches exactly with the same annotated part).
In CUB and PartImageNet we report, in addition, the classification score on the same test set used for part quality evaluation. In the case of CelebA, the classes on the test set do not overlap with those in the training set.


\myparagraph{Implementation details}
We trained all our models with Adam, with a starting learning rate of $10^{-4}$ for the ResNet-101 backbone, $10^{-3}$ for the new layers, and $10^{-2}$ for the modulation vectors. We apply 5 reductions by 0.5 every 5 epochs for CUB and PartImageNet and every 3 for CelebA. The loss weights were all set to 1 except for $\mathcal{L}_\text{conc}$, where a weight of 1000 was used because of its much lower magnitude. This setting was decided based on the results on the CUB validation set and kept constant on all experiments afterwards. For~\cite{huang_interpretable_2020}, we used $\alpha=1$ on CUB and CelebA and $\alpha=0.002$ on PartImageNet.


\subsection{Quantitative results}

The results in Table~\ref{tab:quant} show that, on CUB with $K=4$ parts, our method already performs comparably to~\cite{choudhury_unsupervised_2022}, with 9.12\% keypoint regression error vs. 9.20\% in~\cite{choudhury_unsupervised_2022}, even though~\cite{choudhury_unsupervised_2022} use spatially explicit foreground masks at train time. Our method obtains better results than all other methods in all settings and on all metrics, improving over the second best method~\cite{amir_deep_2022_dinovit} from 50.57 to 56.87 NMI and 26.14 to 38.05 ARI for $K=16$, all while improving the classification accuracy over~\cite{huang_interpretable_2020} and a ResNet101 trained in the same setting. Interestingly, increasing the number of parts not only results in a substantial improvement on the part quality metrics, but also in classification accuracy, from 86.17\% with $K=4$ to 87.49\% with $K=16$, unlike for~\cite{huang_interpretable_2020}, with which the classification accuracy is reduced as the number of parts increases.

On CelebA, our method obtains the best clustering scores on all settings, improving for $K=4$ over~\cite{huang_interpretable_2020} from 56.69 to 75.97 NMI and from 34.74 to 69.53 ARI, thus doubling the result of the best competing method. However,~\cite{huang_interpretable_2020} does result in lower keypoint regression errors. We also obtain better keypoint regression errors than~\cite{amir_deep_2022_dinovit}, 11.11\% vs. 11.36\%, although this method completely fails when evaluated in terms of the clustering metrics. As can be seen in Section~\ref{sec:qual}, this is related to the fact that this method is task agnostic and focuses on elements not related to facial landmarks, such as clothing and hair, which are not as useful for locating facial landmarks. With a single part being assigned to the face, Dino ViT~\cite{amir_deep_2022_dinovit} obtains much lower clustering scores than the other methods.

In the case of PartImageNet, a more challenging dataset in terms of class diversity, Table~\ref{tab:quant} shows that both our method and Dino ViT~\cite{amir_deep_2022_dinovit} are competitive, with our method taking the lead in terms of NMI: 41.49 with PDiscoNet vs. 37.81 with Dino ViT, and Dino ViT in terms of ARI: 14.17 with PDiscoNet and 16.50 with Dino ViT. The method by Huang and Li \cite{huang_interpretable_2020} fails to capture the diversity in terms of part semantics, resulting in very low NMI and ARI, 10.19 and 1.05, and much lower classification scores, with a maximum of 74.22\% with $K=4$, while our model reaches 89.00\% with $K=25$, close to the 90.81\% obtained by ResNet101 in the same training settings.

\begin{table*}[t]
\centering
\begin{tabular}{l|ccc|c|| cc|c}
 & \multicolumn{4}{c||}{\textbf{CUB}} & \multicolumn{3}{c}{\textbf{PartImageNet}}\\
                & Kp. $\downarrow$ & NMI $\uparrow$ & ARI $\uparrow$ & Class. $\uparrow$ & NMI $\uparrow$ & ARI $\uparrow$ & Class. $\uparrow$  \\
                \hline\hline
Full model                          & \textbf{7.67}  &   \textbf{56.87} & \textbf{38.05} & \textbf{87.49} &   \textbf{27.13} & \textbf{8.76} & 88.58\\ \hline
No $\mathcal{L}_\text{orth}$               & 10.29 & 36.12 & 19.41 & 86.17 & 16.25 & 4.59 & 89.12 \\
No $\mathcal{L}_\text{equiv}$                 & 10.31 & 40.22 & 21.32 & 86.60  & 17.55 & 4.22 & 89.90  \\
No $\mathcal{L}_\text{pres}$                        & 7.72  & 55.18 & 35.69 & 87.21  & 12.22 & 3.84 & 88.52 \\
No $\mathcal{L}_\text{conc}$               & 8.58  & 52.44 & 32.17 & 86.77  & 19.71 & 7.39 & \textbf{90.32} \\
No modulation                       & 8.05  & 53.45 & 35.90 & 86.36  & 19.83 & 6.02 & 89.42 \\
No part dropout                 & 8.48  & 46.37 & 25.36 & 86.93  & 19.97 & 4.65 & 89.72 \\
\end{tabular}
\caption{Ablation studies on CUB with $K=16$ and PartImageNet with $K=8$.}
\label{tab:abl}
\end{table*}

\subsection{Ablation studies}

The ablation results in Table~\ref{tab:abl} confirm that all ingredients in the method are important to obtain competitive results on part discovery. On CUB, $\mathcal{L}_\text{orth}$, $\mathcal{L}_\text{equiv}$ and part dropout seem to individually contribute the most to part quality in terms of the clustering metrics, while $\mathcal{L}_\text{conc}$ seems to play an important role to improve the keypoint regression results. On the other hand, $\mathcal{L}_\text{orth}$ and part feature vector modulation are the elements with the highest impact on classification performance, which on CUB is positively impacted by all terms. $\mathcal{L}_\text{pres}$ seems to only have a very marginal impact on both part discovery and classification performance on CUB. However, on PartImageNet it is the most important of all terms and the only one that does not hurt the classification accuracy. This is likely to stem from the very different distribution of parts in each dataset. On CUB, on the one hand, all parts are shared by all objects in the dataset, since they are all birds, and a majority of them is visible in all images. On PartImagenet, on the other hand, the different categories of classes do not naturally share the exact same parts (e.g. snakes, vehicles and birds). Forcing all parts to be present in each batch would prevent one single part prototype from dominating and becoming an object detector rather than a part detector, as happens in PartImageNet with the method of~\cite{huang_interpretable_2020}, as seen in Fig.~\ref{fig:qual_pim}.

\subsection{Sensitivity studies}

We have performed a part-dropout rate sensitivity analysis, shown in Table~\ref{tab:sensitivity}. In general, increasing the dropout rate improves the part discovery performance, with the best results obtained with the highest tested rate of 0.9, with NMI going up to 54.42 from 49.22 when a part dropout rate of 0.3 is used. 
Such a high rate means that every part needs to capture enough information to be able to classify the image, since very often all parts except one are dropped-out. 
This poses a very strict constraint on the part discovery process that prevents the appearance of spurious part prototypes.
However, this tends to negatively impact the classification performance, which drops from 87.31\% to 83.34\%, probably due to the fact that, by trying to learn parts that are able to perform classification on their own, there is a lack of incentives for the model to learn the complementarities between parts.
A value between 0.3 and 0.7 provides a good compromise between the two tasks.

\begin{table}[h]
\centering
{\begin{tabular}{l|ccccc}
Dropout rate             & 0.1 & 0.3 & 0.5 & 0.7 & 0.9 \\
                \hline\hline
NMI $\uparrow$                            &   45.30 & 49.22 & 50.08 & 49.63 & 54.42 \\ 
ARI $\uparrow$             & 22.78 & 27.27 & 26.96 & 29.15 & 34.56 \\ \hline
Class. $\uparrow$                  & 86.90 & 87.31 & 86.72 & 86.26 & 83.34  \\
\end{tabular}}
\caption{Part-dropout rate sensitivity on CUB with $K=8$.}
\label{tab:sensitivity}
\end{table}

We also investigate the behaviour of our method with respect to the presence of noise at test time. The results in Table~\ref{tab:noise} show that the classification accuracy of our method is higher than the most closely related method \cite{huang_interpretable_2020} when the input images are subjected to Gaussian noise. Apart from the absolute accuracy of our method being higher, the percentual decrease between each increase in noise is lower, suggesting that an improved ability in part localization also carries advantages in terms of robustness to noise.

\begin{table}[h]
\centering
{\begin{tabular}{l|cccccc}
Noise SD             & 0.03 & 0.07 & 0.15 & 0.30 & 0.75 & 1.50 \\
                \hline\hline
Huang \cite{huang_interpretable_2020}   & 84.2
& 82.1
& 78.2
& 71.4
& 50.5
& 22.1

 \\ 
PDiscoNet          &  85.6
& 85.6
& 82.2
& 77.9
& 63.3
& 38.2
\end{tabular}}
\caption{Class. acc. with Gaussian noise on CUB with $K=16$.}
\label{tab:noise}
\end{table}

\subsection{Qualitative results}
\label{sec:qual}

\begin{figure}[t]
\centering
\setlength\tabcolsep{1.5pt} 
 \begin{tabular}{ccccc}
  Image & Dino \cite{amir_deep_2022_dinovit}  & \cite{huang_interpretable_2020} (all) & \cite{huang_interpretable_2020} $>0.1$ & PDiscoNet\\
  \includegraphics[width=1.6 cm]{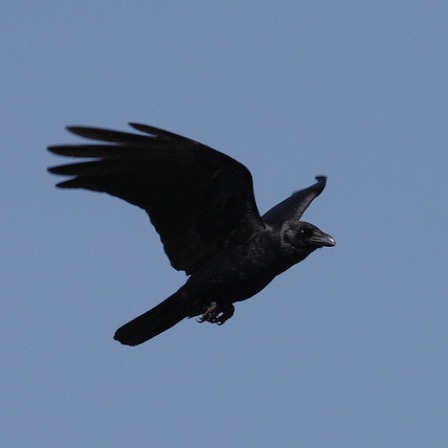} &
  \includegraphics[width=1.6 cm, height=1.6 cm]{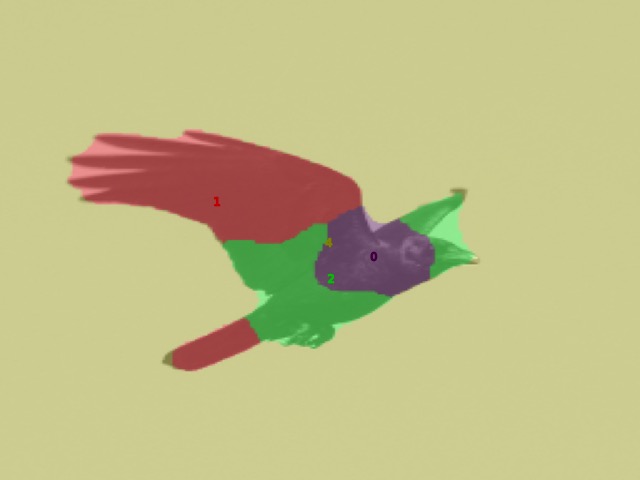}        &
  \includegraphics[width=1.6 cm, height=1.6 cm]{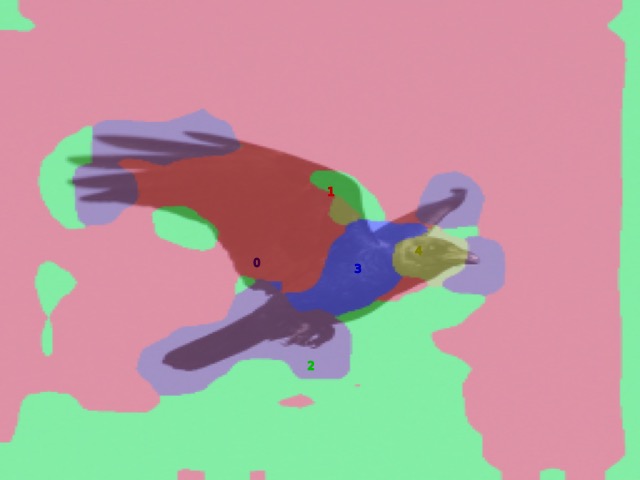}    &
  \includegraphics[width=1.6 cm, height=1.6 cm]{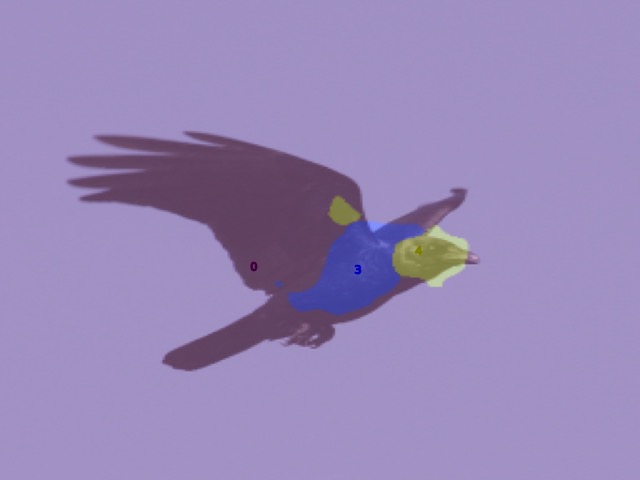}    &
  \includegraphics[width=1.6 cm, height=1.6 cm]{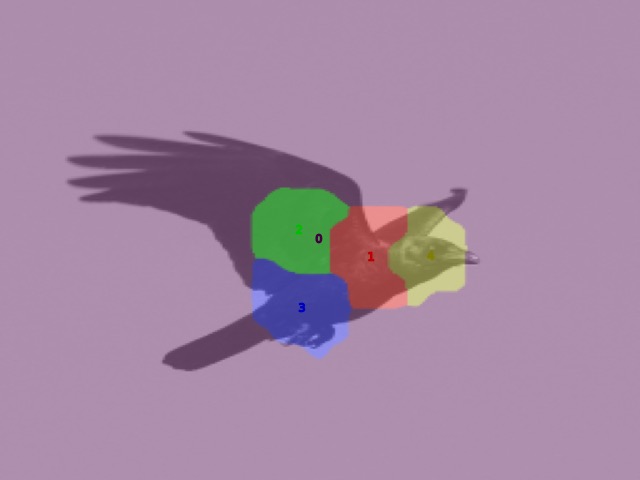}\\
  \includegraphics[width=1.6 cm]{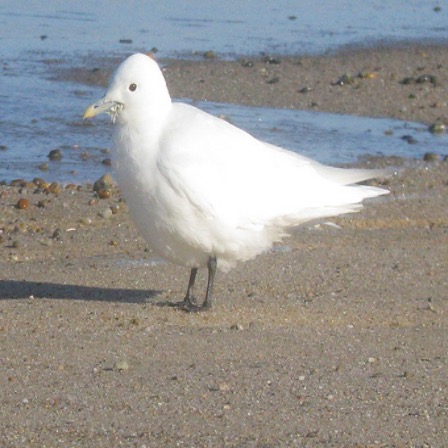} &
  \includegraphics[width=1.6 cm, height=1.6 cm]{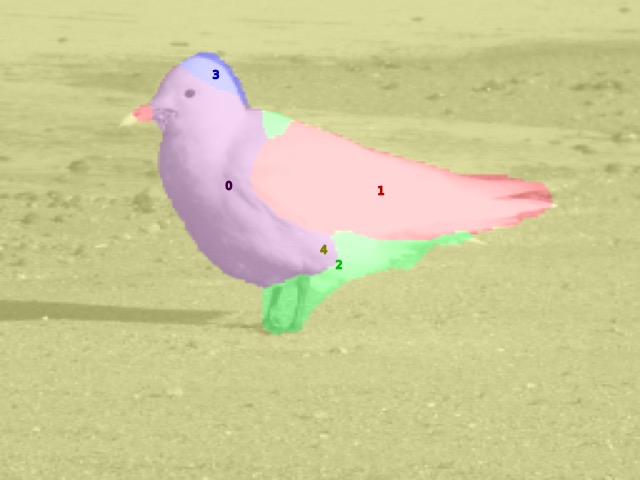}        &
  \includegraphics[width=1.6 cm, height=1.6 cm]{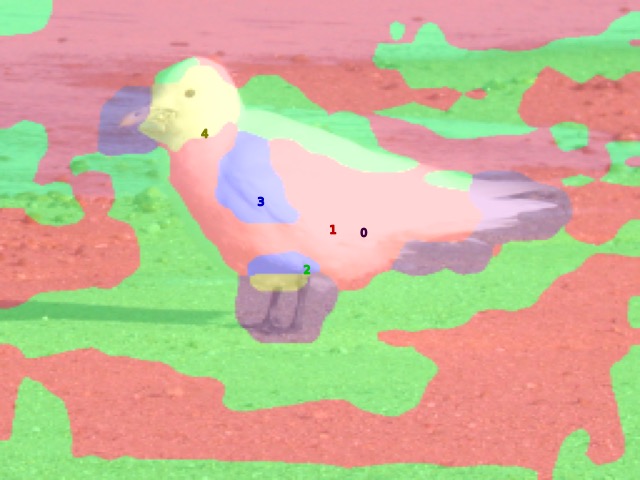}        &
  \includegraphics[width=1.6 cm, height=1.6 cm]{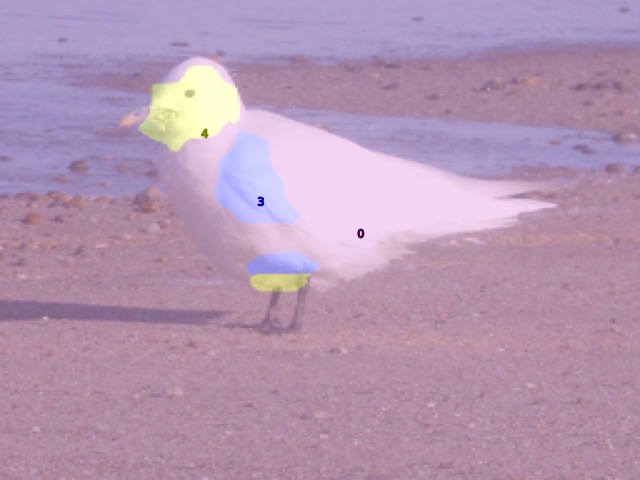}        &
  \includegraphics[width=1.6 cm, height=1.6 cm]{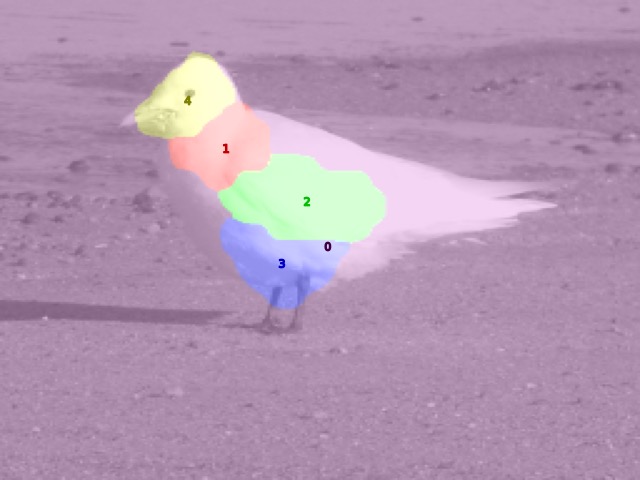}\\
  \vspace{0.25 cm}
  \includegraphics[width=1.6 cm]{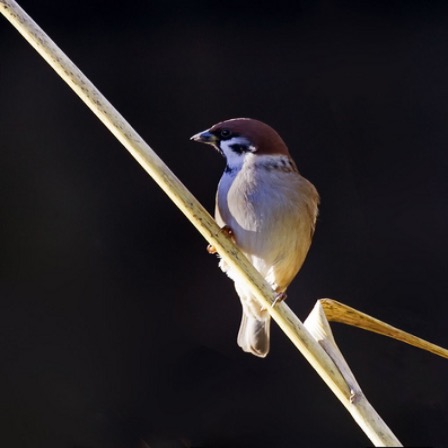} &
  \includegraphics[width=1.6 cm, height=1.6 cm]{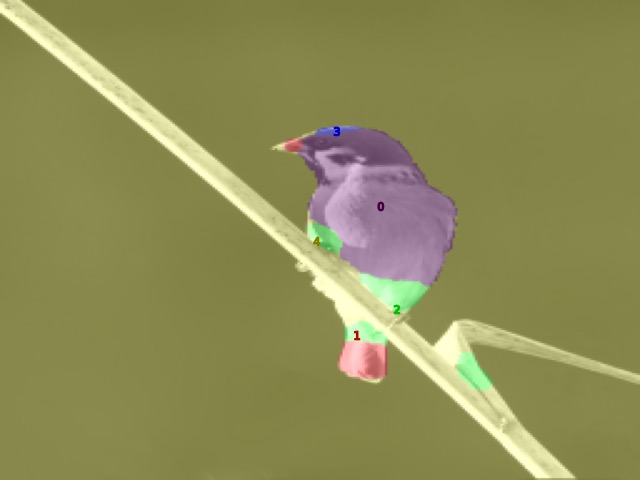}        &
  \includegraphics[width=1.6 cm, height=1.6 cm]{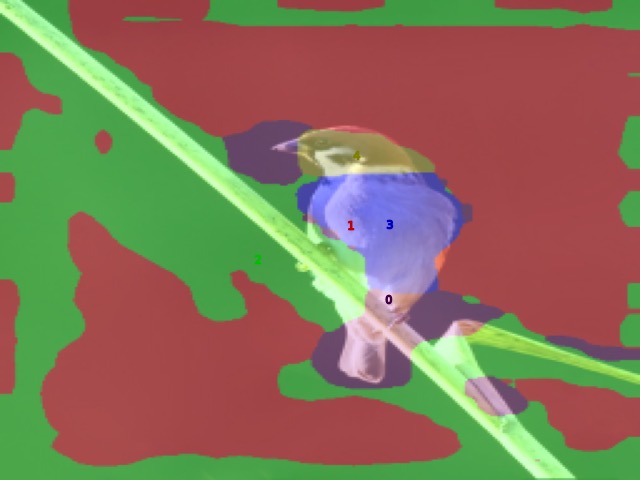}        &
  \includegraphics[width=1.6 cm, height=1.6 cm]{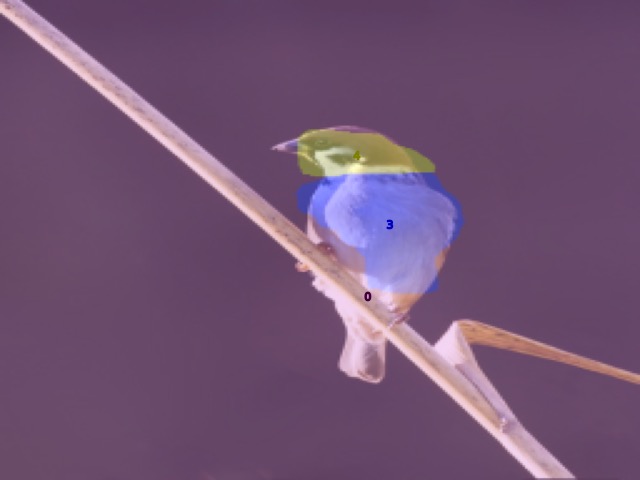}        &
  \includegraphics[width=1.6 cm, height=1.6 cm]{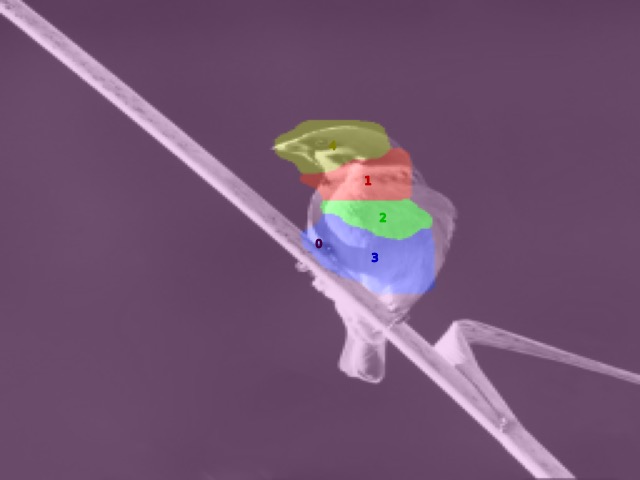}\\

  \includegraphics[width=1.6 cm]{figures_1/cub/original/originalfishcrow.jpg} &
  \includegraphics[width=1.6 cm, height=1.6 cm]{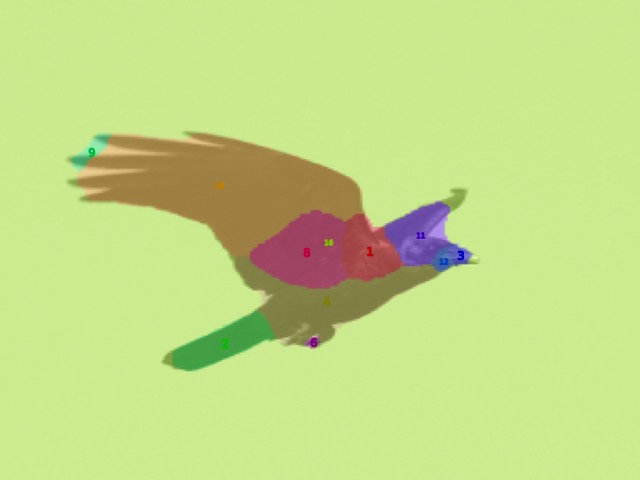}        &
  \includegraphics[width=1.6 cm, height=1.6 cm]{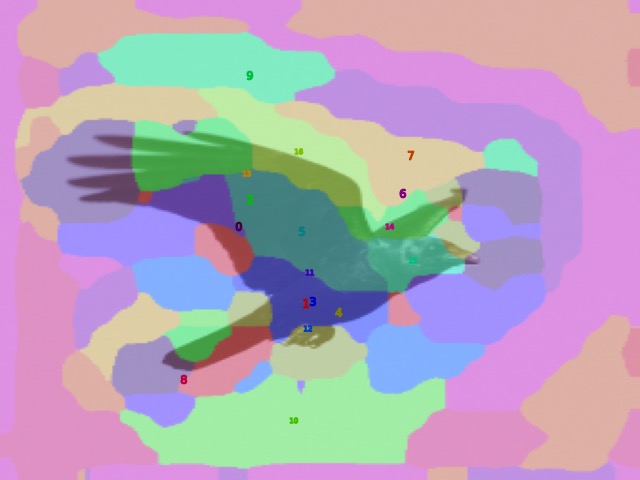}    &
  \includegraphics[width=1.6 cm, height=1.6 cm]{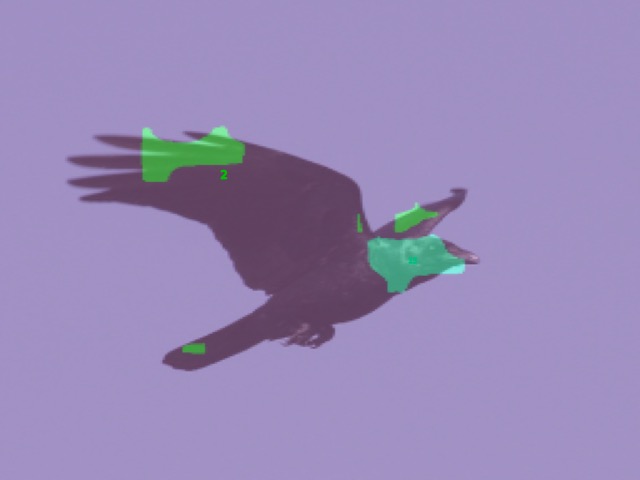}    &
  \includegraphics[width=1.6 cm, height=1.6 cm]{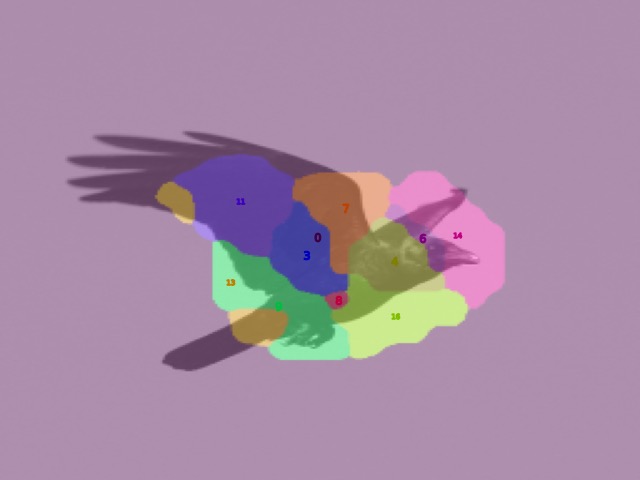}\\
  \includegraphics[width=1.6 cm]{figures_1/cub/original/originalgull.jpg} &
  \includegraphics[width=1.6 cm, height=1.6 cm]{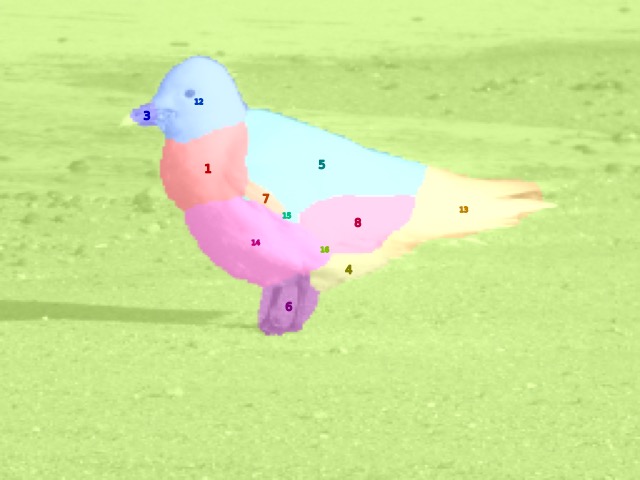}        &
  \includegraphics[width=1.6 cm, height=1.6 cm]{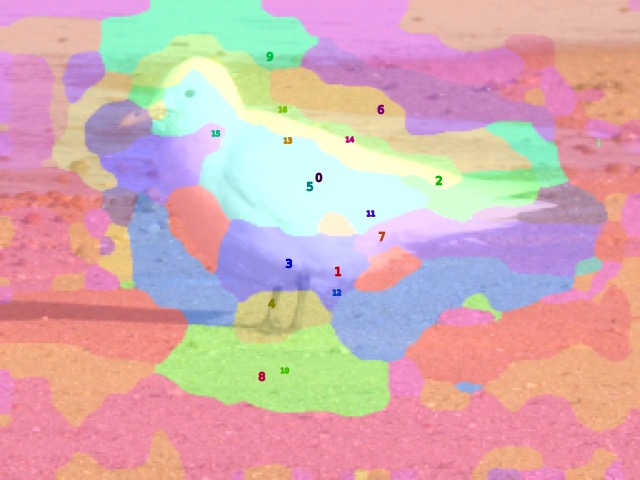}        &
  \includegraphics[width=1.6 cm, height=1.6 cm]{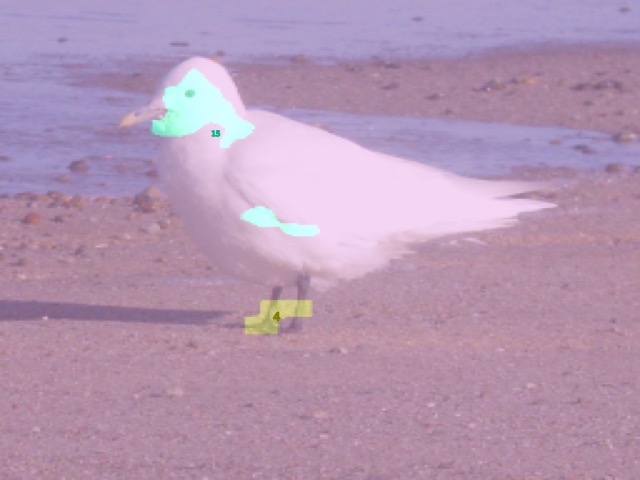}        &
  \includegraphics[width=1.6 cm, height=1.6 cm]{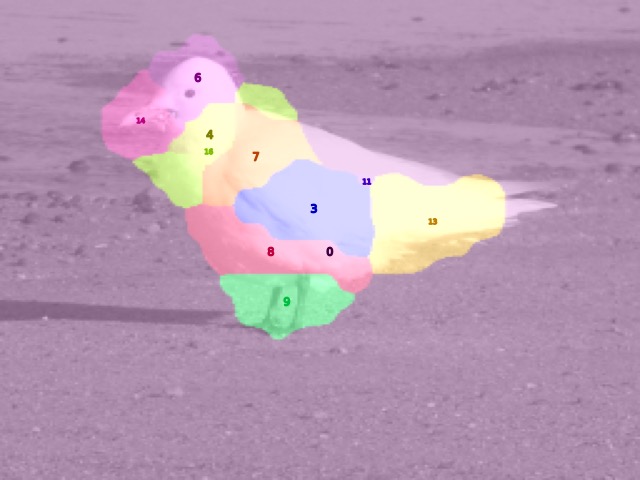}\\
  \includegraphics[width=1.6 cm]{figures_1/cub/original/originaltreesparrow.jpg} &
  \includegraphics[width=1.6 cm, height=1.6 cm]{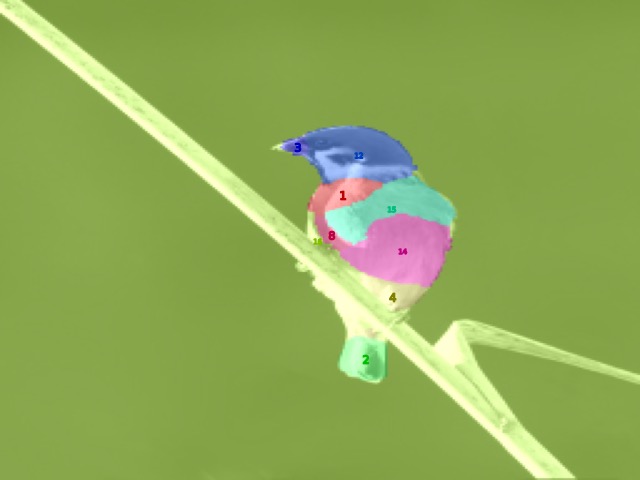}        &
  \includegraphics[width=1.6 cm, height=1.6 cm]{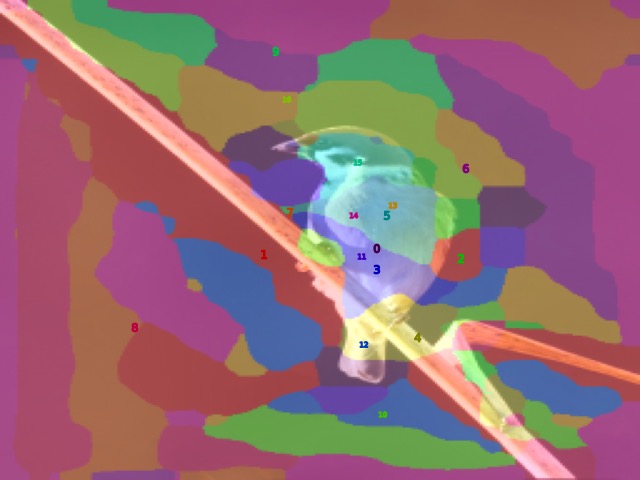}        &
  \includegraphics[width=1.6 cm, height=1.6 cm]{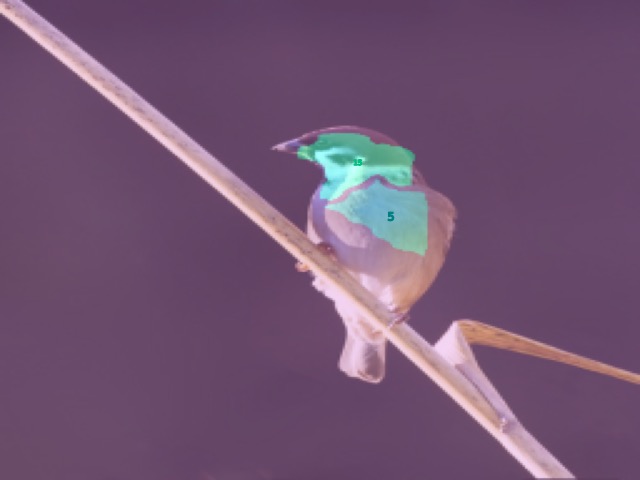}        &
  \includegraphics[width=1.6 cm, height=1.6 cm]{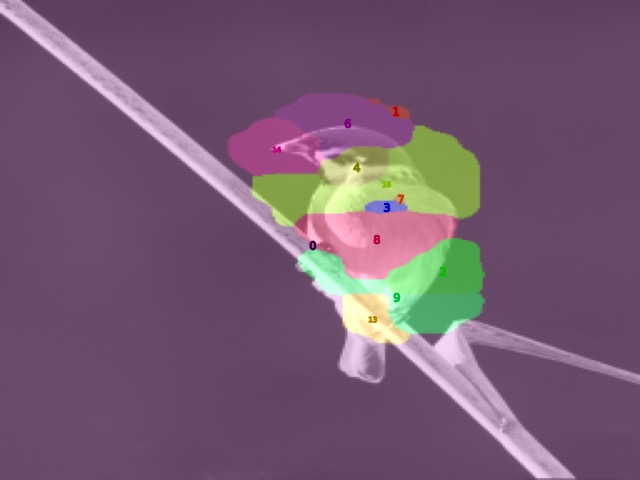}\\
 \end{tabular}
 \caption{Qualitative results on CUB for our method, \cite{huang_interpretable_2020} w/ and w/o part map thresholding, and \cite{amir_deep_2022_dinovit}. Top rows: all methods with $k=4$. Bottom rows: $K=16$.}
 \label{fig:qual_cub}
\end{figure}

\begin{figure}[t]
\centering
\setlength\tabcolsep{1.5pt} 
 \begin{tabular}{ccccc}
  Image & Dino \cite{amir_deep_2022_dinovit}  & \cite{huang_interpretable_2020} all & \cite{huang_interpretable_2020} $>0.1$ & PDiscoNet \\
  \includegraphics[width=1.6 cm]{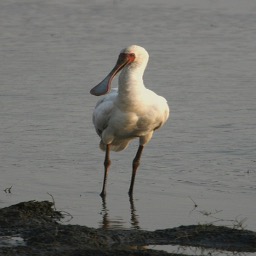} &
  \includegraphics[width=1.6 cm, height=1.6 cm]{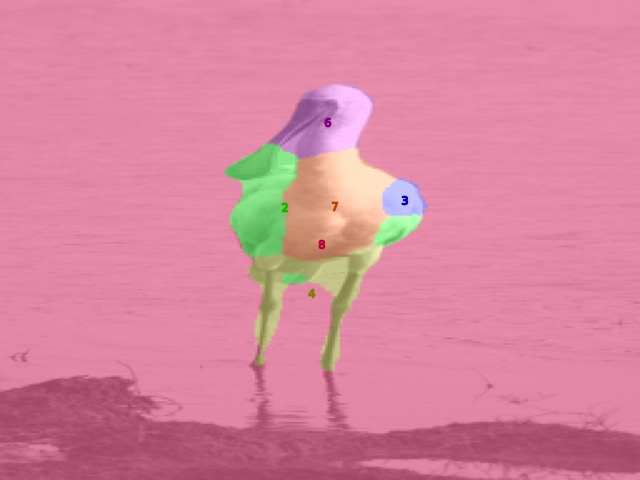}        &
  \includegraphics[width=1.6 cm, height=1.6 cm]{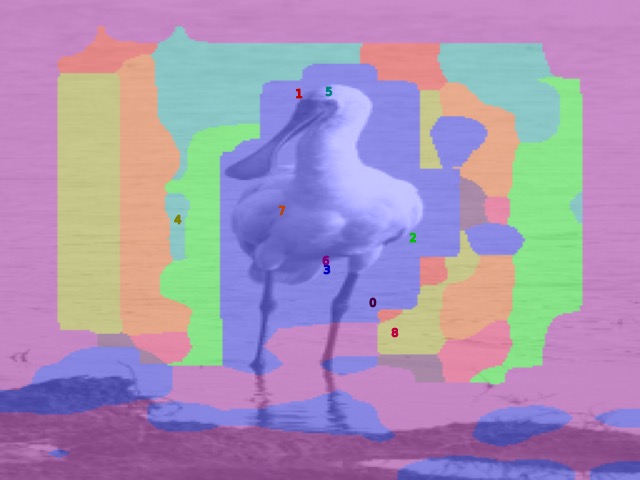}    &
  \includegraphics[width=1.6 cm, height=1.6 cm]{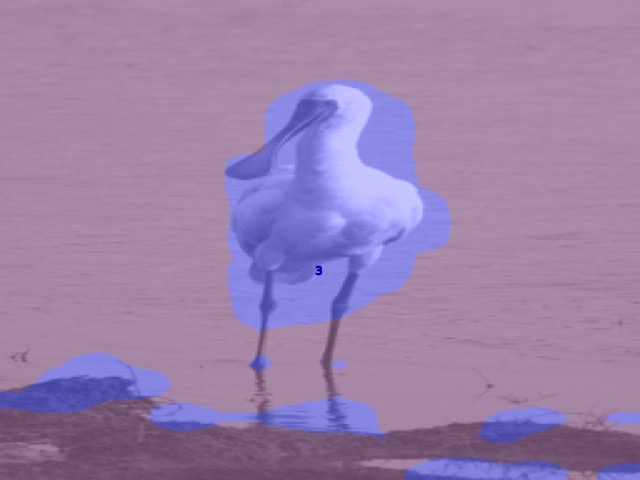}    &
  \includegraphics[width=1.6 cm, height=1.6 cm]{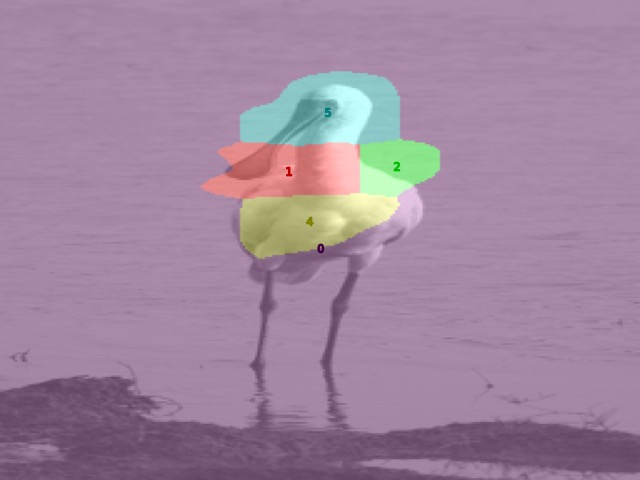}\\
  \includegraphics[width=1.6 cm]{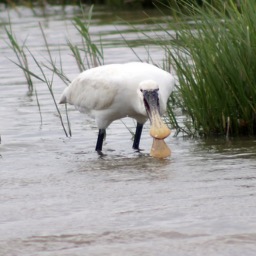} &
  \includegraphics[width=1.6 cm, height=1.6 cm]{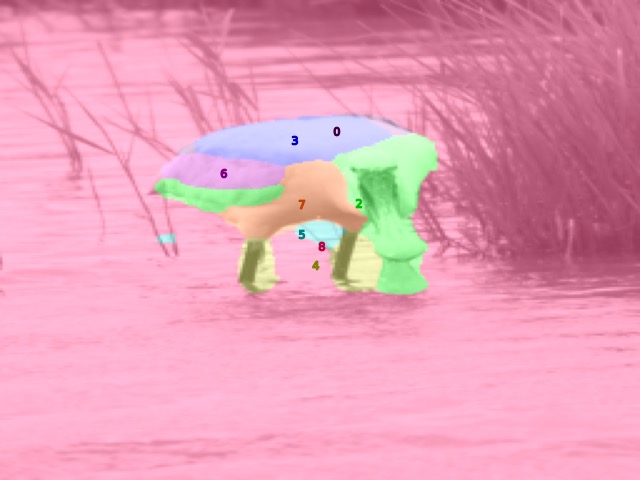}        &
  \includegraphics[width=1.6 cm, height=1.6 cm]{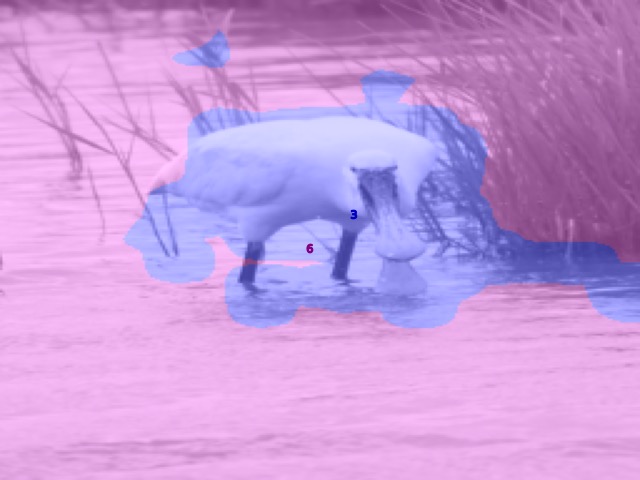}    &
  \includegraphics[width=1.6 cm, height=1.6 cm]{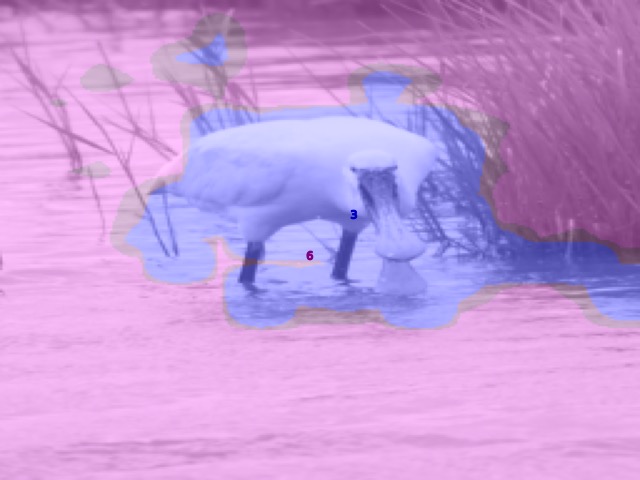}    &
  \includegraphics[width=1.6 cm, height=1.6 cm]{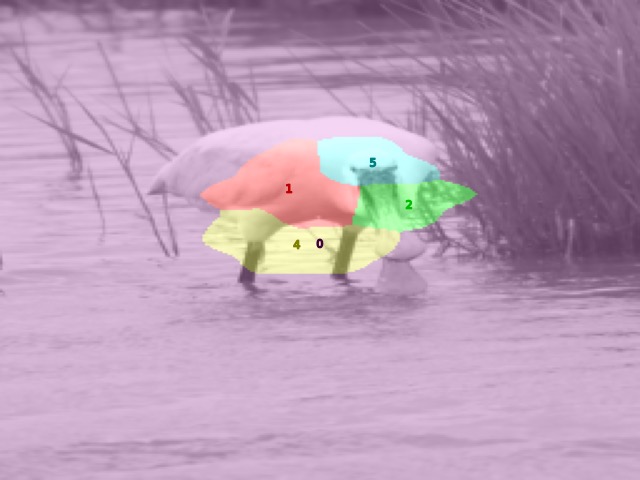}\\
  \vspace{0.25 cm}
  \includegraphics[width=1.6 cm]{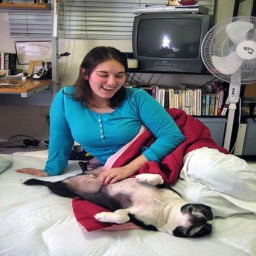} &
  \includegraphics[width=1.6 cm, height=1.6 cm]{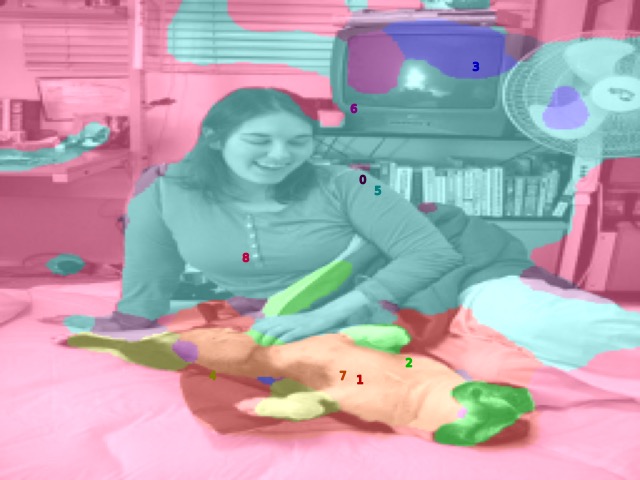}        &
  \includegraphics[width=1.6 cm, height=1.6 cm]{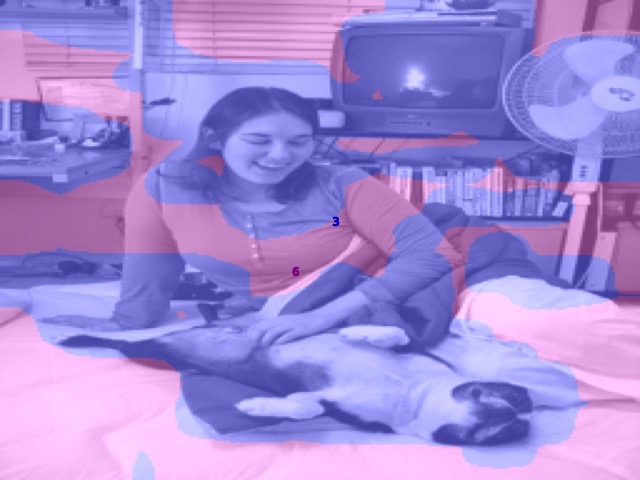}    &
  \includegraphics[width=1.6 cm, height=1.6 cm]{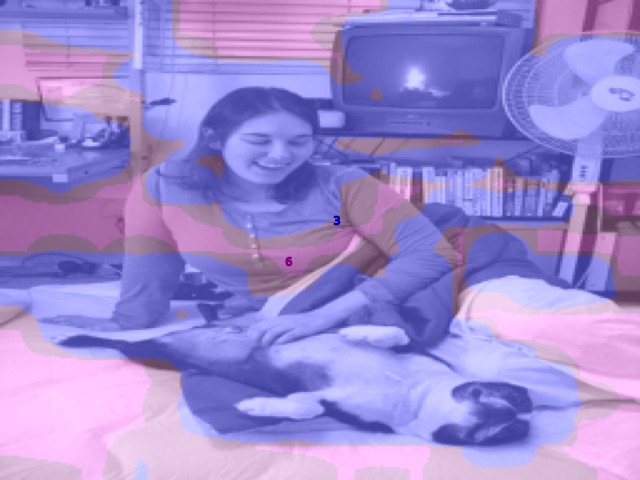}    &
  \includegraphics[width=1.6 cm, height=1.6 cm]{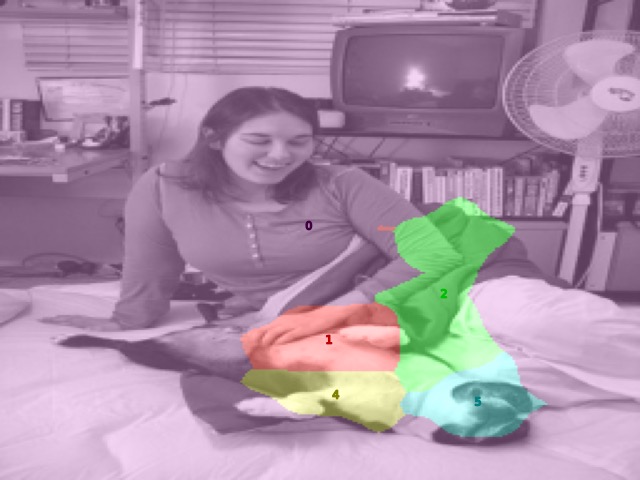}\\

  \includegraphics[width=1.6 cm]{figures_1/partimagenet/original/originalbill1.jpg} &
  \includegraphics[width=1.6 cm, height=1.6 cm]{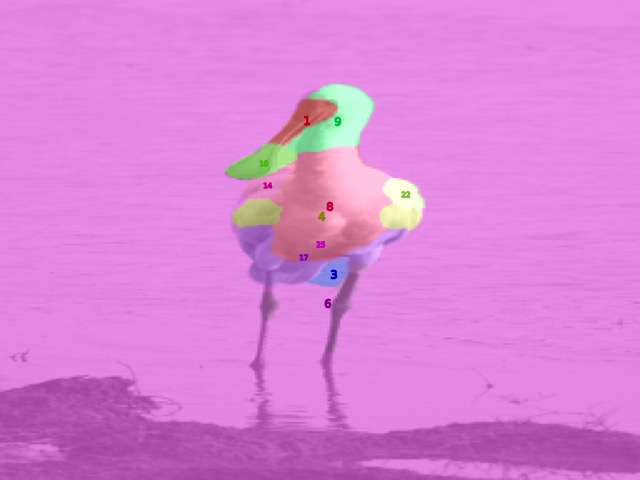}        &
  \includegraphics[width=1.6 cm, height=1.6 cm]{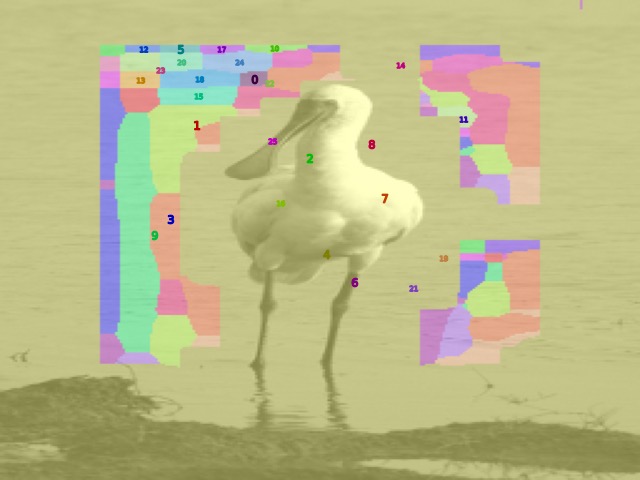}    &
  \includegraphics[width=1.6 cm, height=1.6 cm]{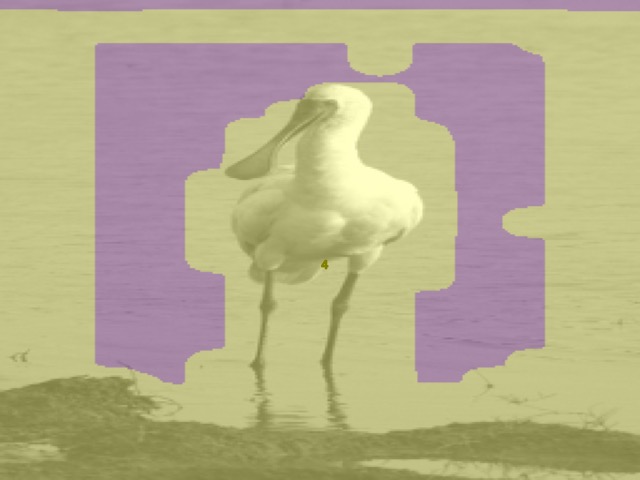}    &
  \includegraphics[width=1.6 cm, height=1.6 cm]{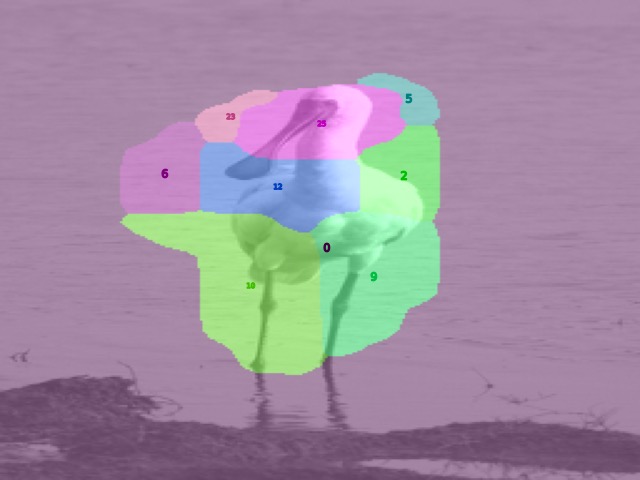}\\
  \includegraphics[width=1.6 cm]{figures_1/partimagenet/original/originalbill2.jpg} &
  \includegraphics[width=1.6 cm, height=1.6 cm]{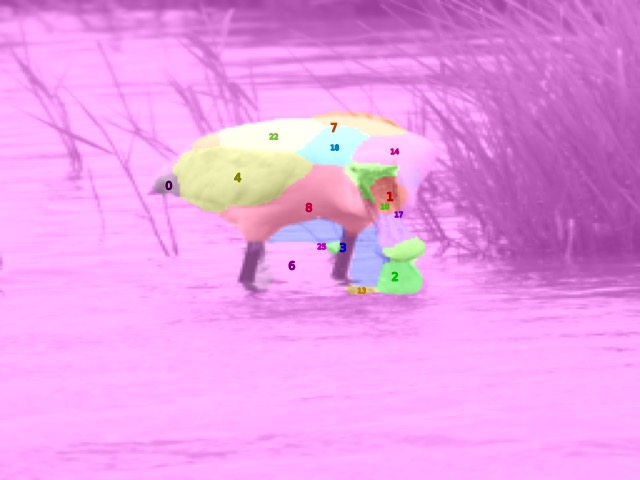}        &
  \includegraphics[width=1.6 cm, height=1.6 cm]{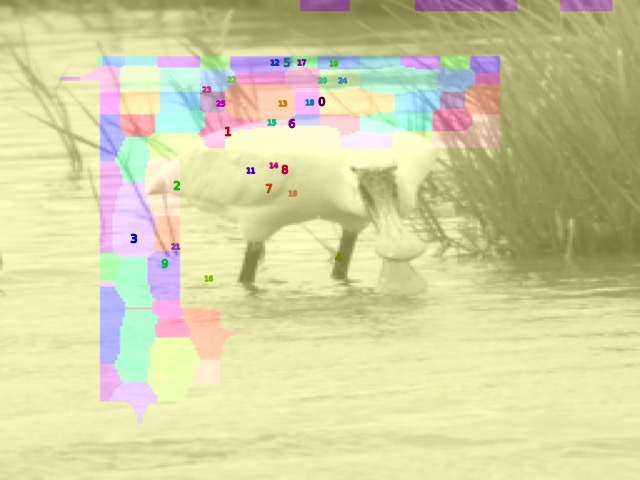}    &
  \includegraphics[width=1.6 cm, height=1.6 cm]{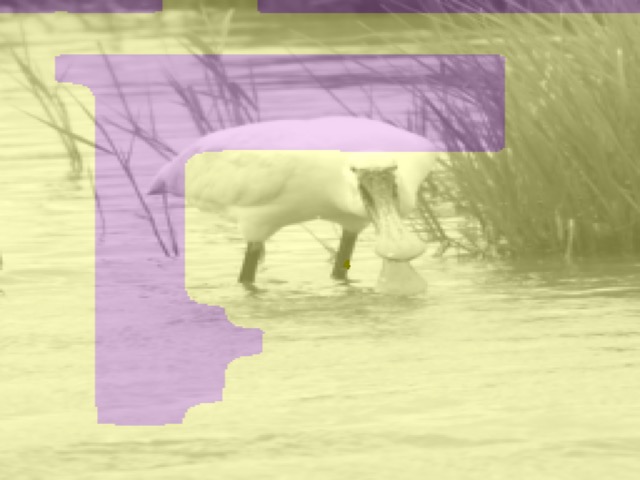}    &
  \includegraphics[width=1.6 cm, height=1.6 cm]{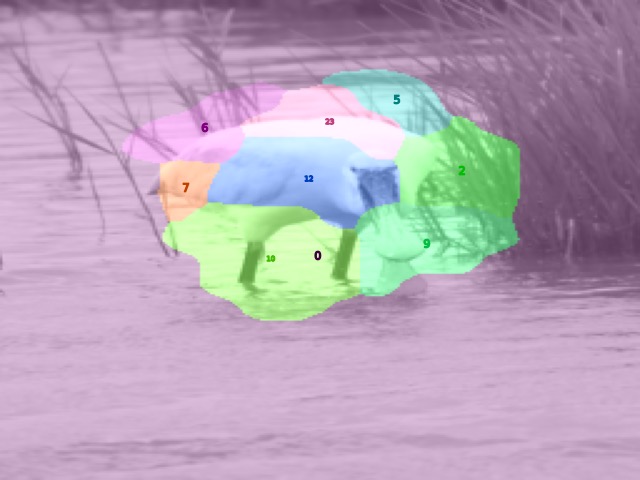}\\\
  \includegraphics[width=1.6 cm]{figures_1/partimagenet/original/originaldog.jpg} &
  \includegraphics[width=1.6 cm, height=1.6 cm]{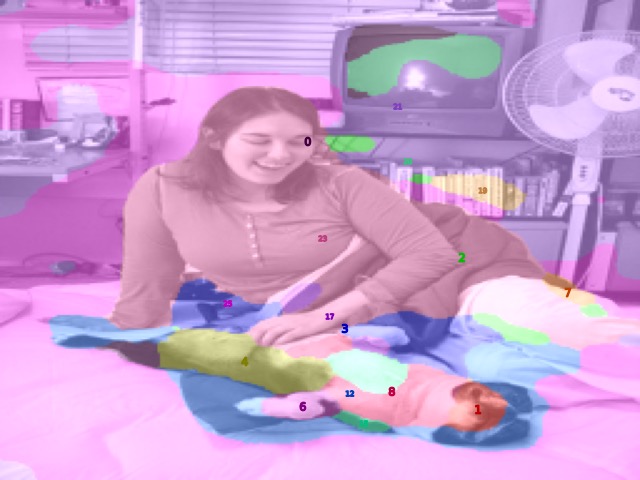}        &
  \includegraphics[width=1.6 cm, height=1.6 cm]{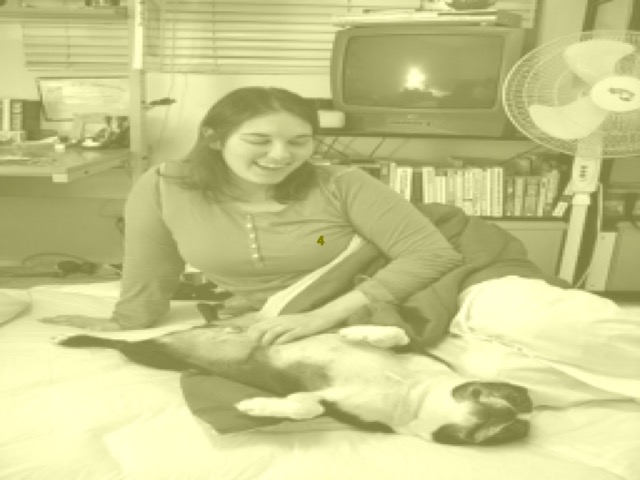}    &
  \includegraphics[width=1.6 cm, height=1.6 cm]{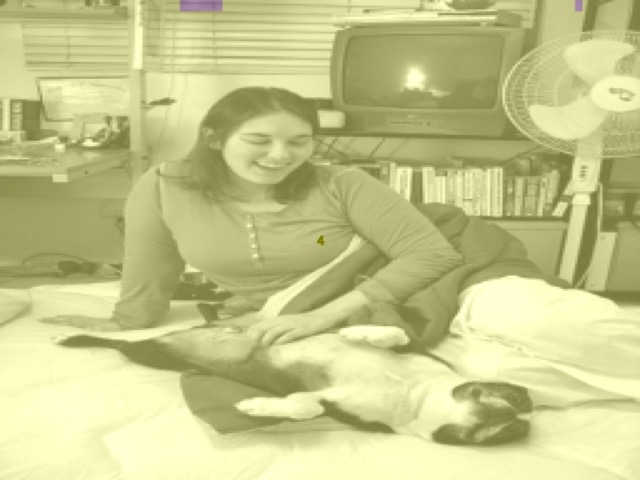}    &
  \includegraphics[width=1.6 cm, height=1.6 cm]{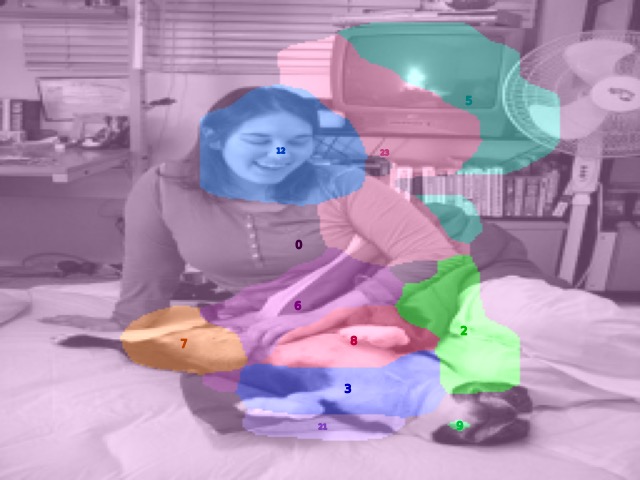}\\
 \end{tabular}
  \caption{Qualitative results on PartImageNet for our method, \cite{huang_interpretable_2020} w/ and w/o part map thresholding, and \cite{amir_deep_2022_dinovit}. Top rows: all methods with $k=8$. Bottom rows: $K=25$.}
  \label{fig:qual_pim}
\end{figure}

In Figs.~\ref{fig:qual_cub} and~\ref{fig:qual_pim} we showcase the effect of increasing the number of parts for the compared methods on CUB ($K=4$ and $K=16$) and PartImageNet ($K=8$ and $K=25$). For the method in~\cite{huang_interpretable_2020}, we show the part assignment maps for all parts and for those with an assigned attention value higher than 0.1, in order to highlight only the image regoins that contribute substantially towards the classification output.
We can see that~\cite{amir_deep_2022_dinovit} and PDiscoNet are able to correctly assign most parts to the foreground objects in the shown examples, even with the increased number of parts (bottom three rows), with~\cite{amir_deep_2022_dinovit} resulting in the best adherence to object boundaries. \cite{huang_interpretable_2020}, on the other hand, assigns only a few parts to the foreground objects, even when more parts are available. In the third column (\cite{huang_interpretable_2020} (all)), we see how the rest of parts are assigned to the background in ways that do not follow any of the objects in the image. When showing only parts that are actually attended to by the classifier (\cite{huang_interpretable_2020} $>0.1$), we confirm that only two or three parts are used in CUB (Fig.~\ref{fig:qual_cub}) with both $K=4$ and $K=16$. 
In the case of PartImageNet (Fig.~\ref{fig:qual_pim}), \cite{huang_interpretable_2020} assigns one single part to the foreground object with $K=8$ and none with $K=16$, while again both~\cite{amir_deep_2022_dinovit} and PDiscoNet tend to assign a diverse set of parts to the foreground. Also in this case we observe that Dino ViT~\cite{amir_deep_2022_dinovit} results in better boundary adherence than PDiscoNet, since we did not explicitly add any element towards this objective.

\begin{figure*}
\centering
\setlength\tabcolsep{1.5pt} 
 \begin{tabular}{lcccccccccc}
\rotatebox{90}{Original} &
 \includegraphics[width=1.6 cm, height=1.6 cm]{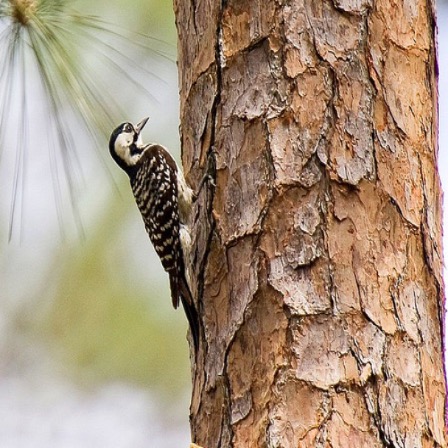} &
 \includegraphics[width=1.6 cm, height=1.6 cm]{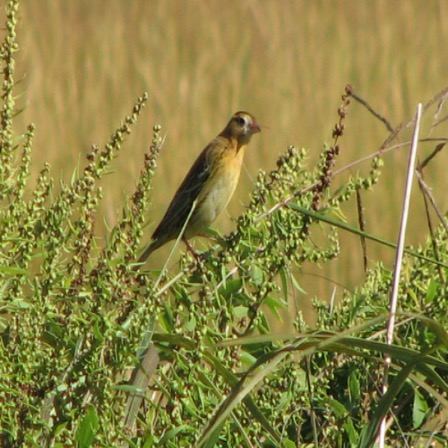}  &
 \includegraphics[width=1.6 cm, height=1.6 cm]{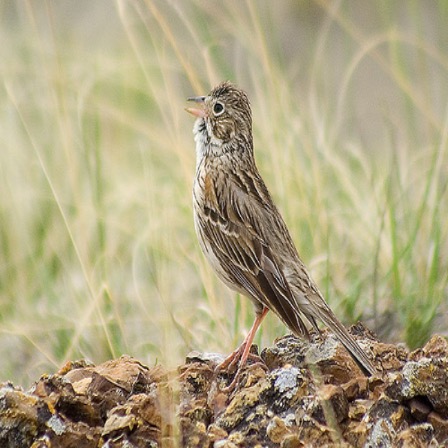}  &
 \includegraphics[width=1.6 cm, height=1.6 cm]{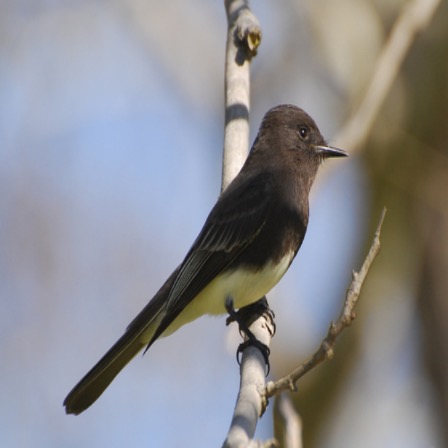}  &
 \includegraphics[width=1.6 cm, height=1.6 cm]{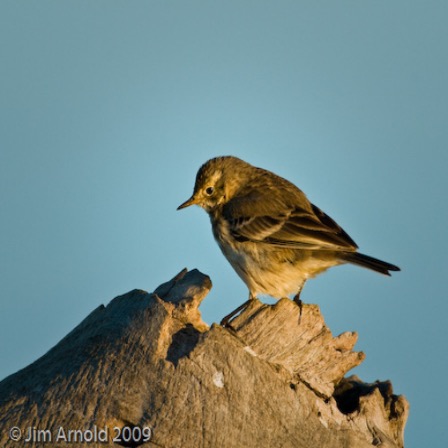}  &
 \includegraphics[width=1.6 cm, height=1.6 cm]{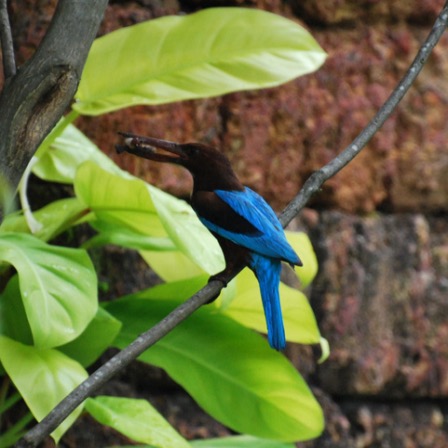}  &
 \includegraphics[width=1.6 cm, height=1.6 cm]{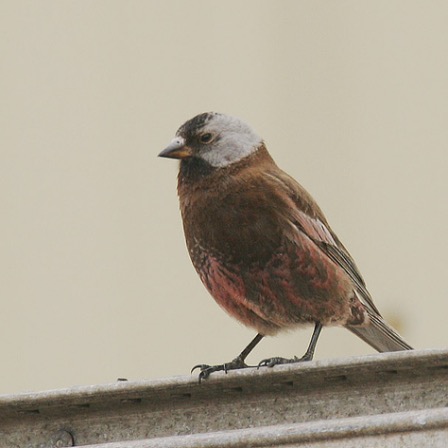}  &
 \includegraphics[width=1.6 cm, height=1.6 cm]{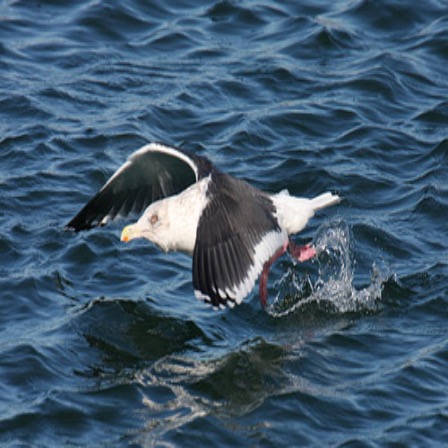}  &
 \includegraphics[width=1.6 cm, height=1.6 cm]{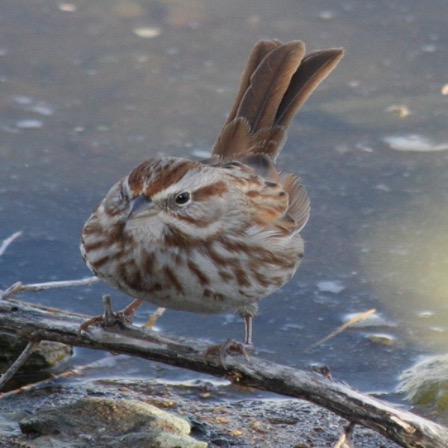}  &
 \includegraphics[width=1.6 cm, height=1.6 cm]{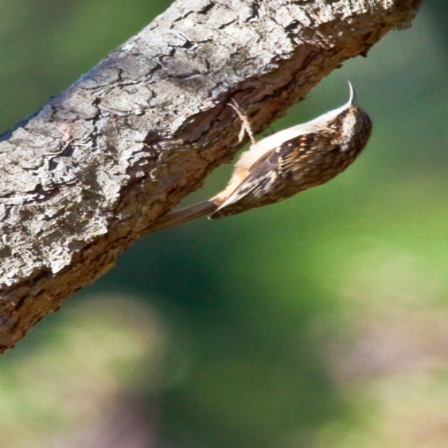}  \\
\rotatebox{90}{Dino~\cite{amir_deep_2022_dinovit}} &
  \includegraphics[width=1.6 cm, height=1.6 cm]{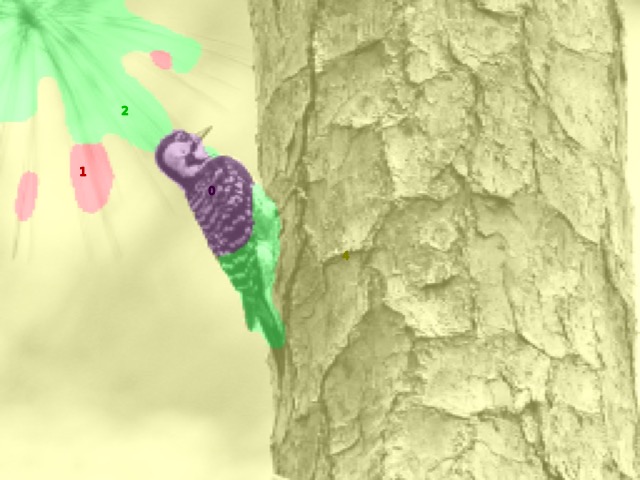} &
 \includegraphics[width=1.6 cm, height=1.6 cm]{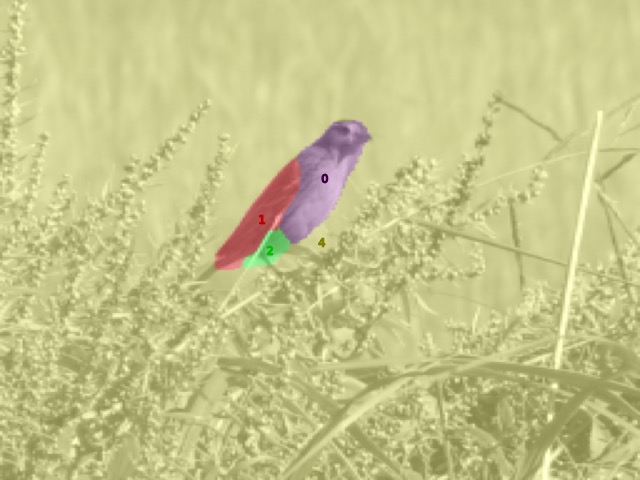}  &
 \includegraphics[width=1.6 cm, height=1.6 cm]{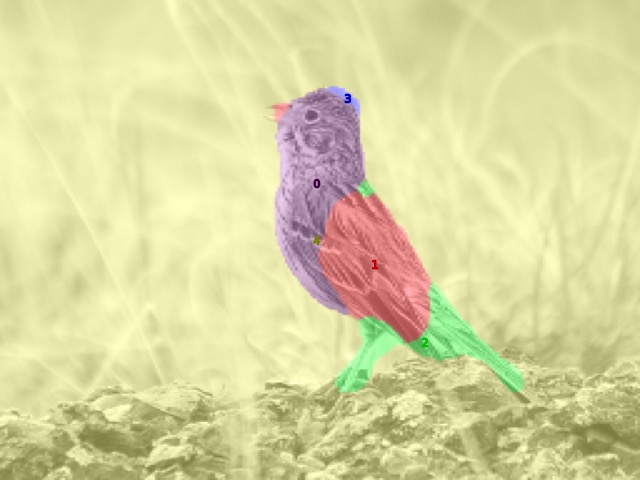}  &
 \includegraphics[width=1.6 cm, height=1.6 cm]{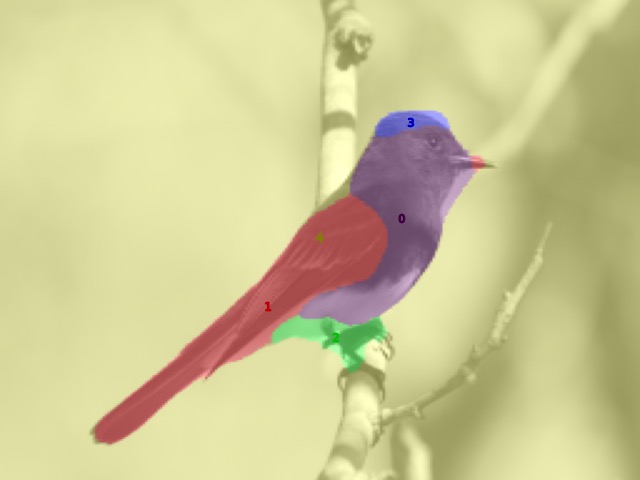}  &
 \includegraphics[width=1.6 cm, height=1.6 cm]{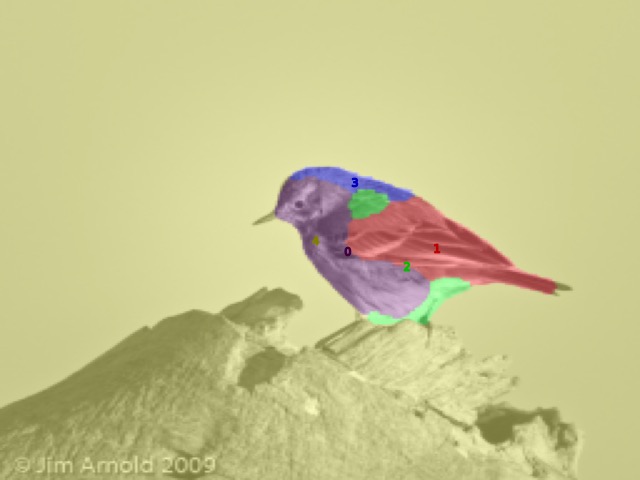}  &
 \includegraphics[width=1.6 cm, height=1.6 cm]{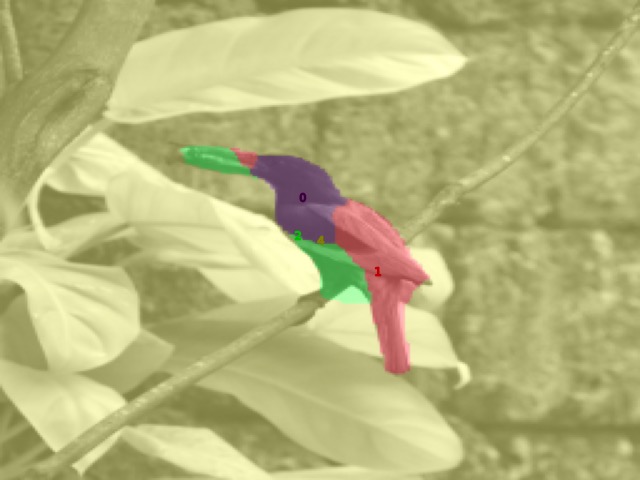}  &
 \includegraphics[width=1.6 cm, height=1.6 cm]{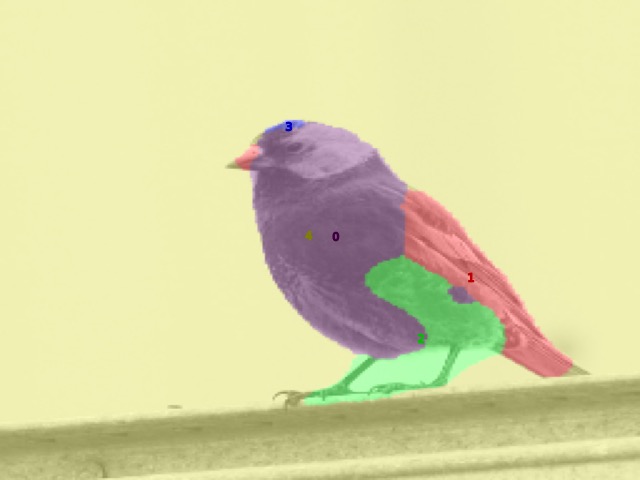}  &
 \includegraphics[width=1.6 cm, height=1.6 cm]{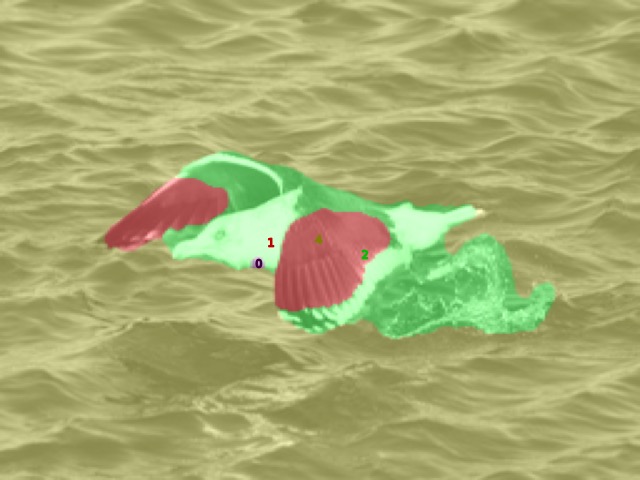}  &
 \includegraphics[width=1.6 cm, height=1.6 cm]{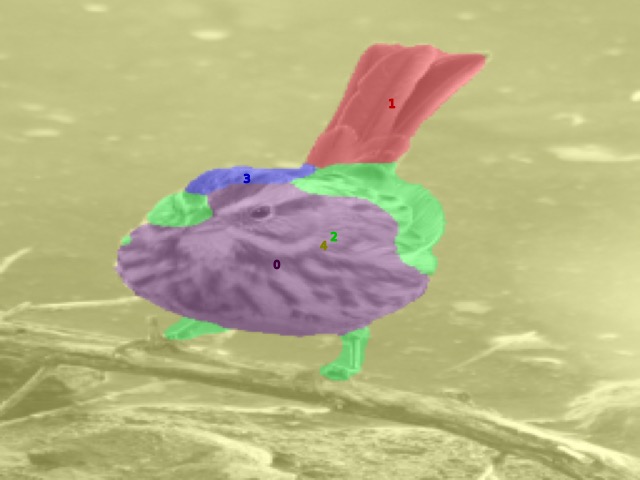}  &
 \includegraphics[width=1.6 cm, height=1.6 cm]{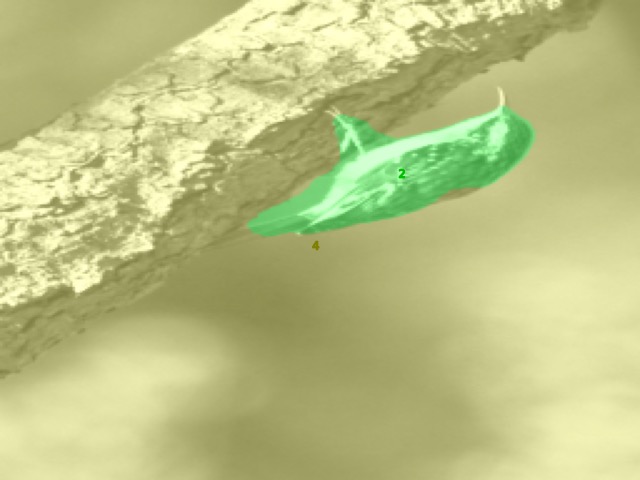} \\

\rotatebox{90}{Huang~\cite{huang_interpretable_2020}} &
 \includegraphics[width=1.6 cm, height=1.6 cm]{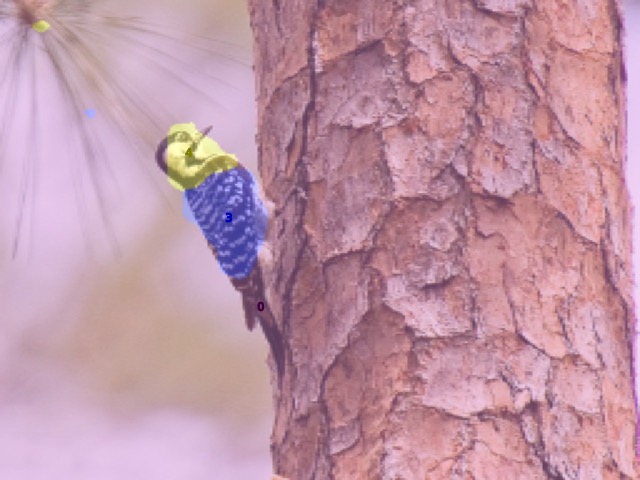} &
 \includegraphics[width=1.6 cm, height=1.6 cm]{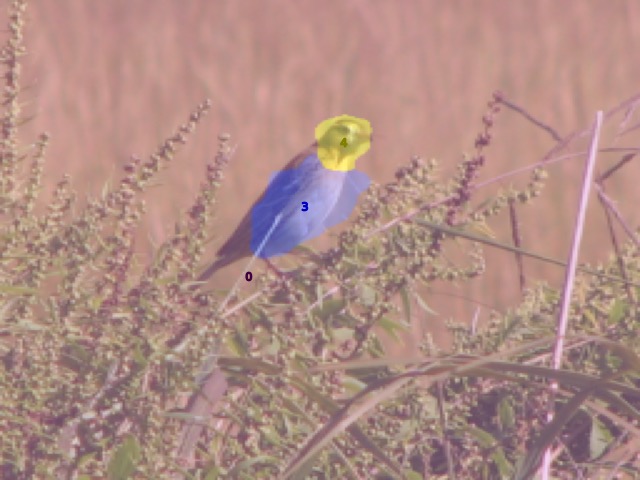}  &
 \includegraphics[width=1.6 cm, height=1.6 cm]{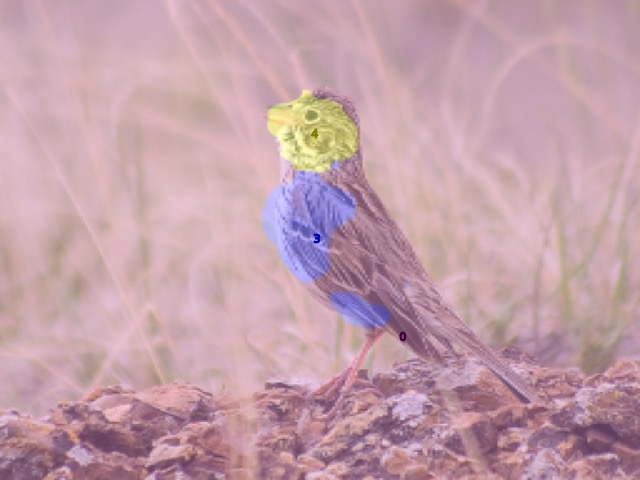}  &
 \includegraphics[width=1.6 cm, height=1.6 cm]{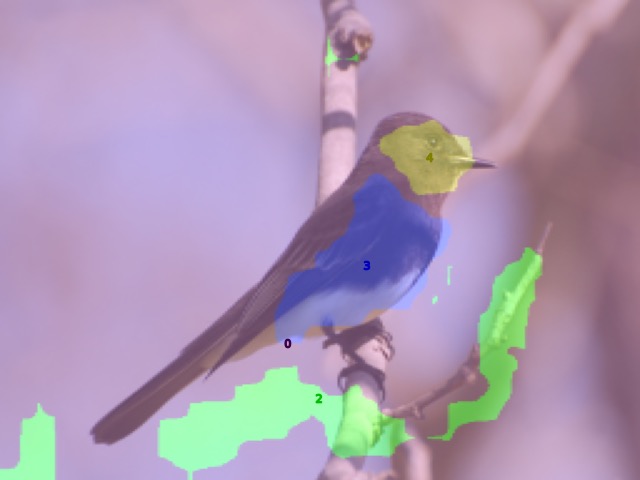}  &
 \includegraphics[width=1.6 cm, height=1.6 cm]{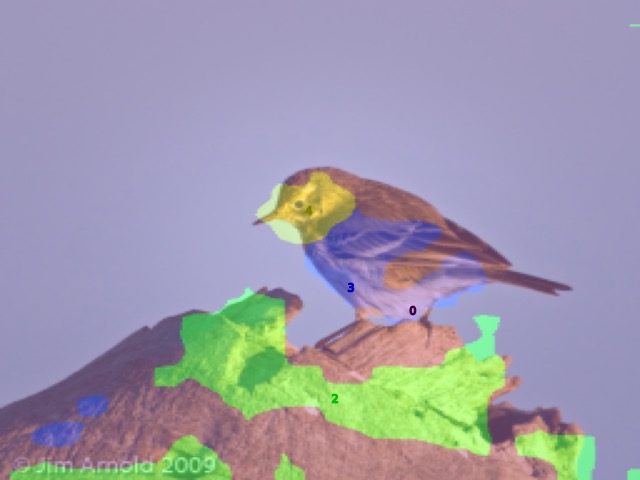}  &
 \includegraphics[width=1.6 cm, height=1.6 cm]{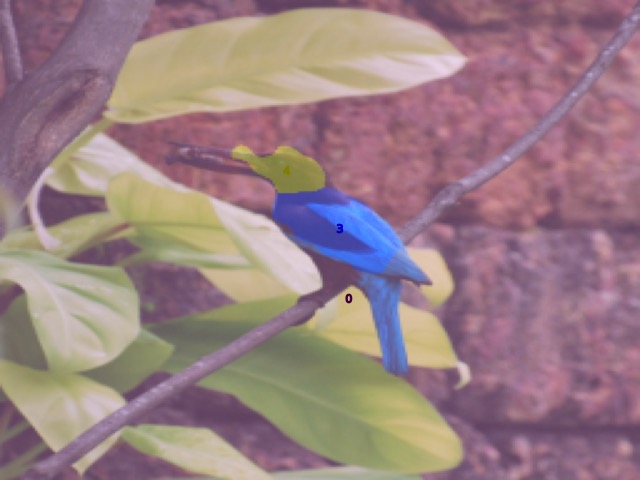}  &
 \includegraphics[width=1.6 cm, height=1.6 cm]{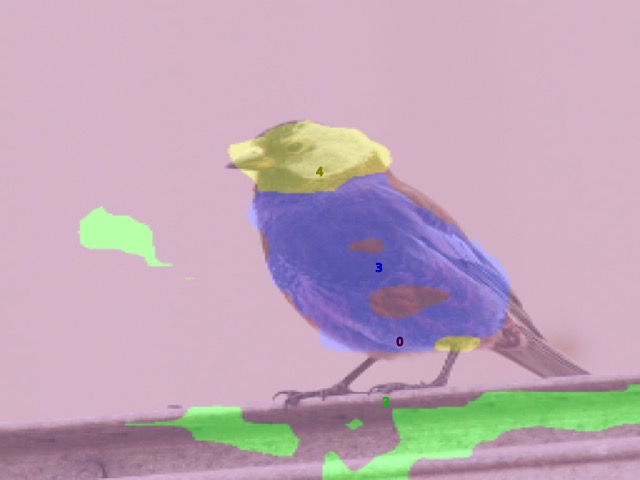}  &
 \includegraphics[width=1.6 cm, height=1.6 cm]{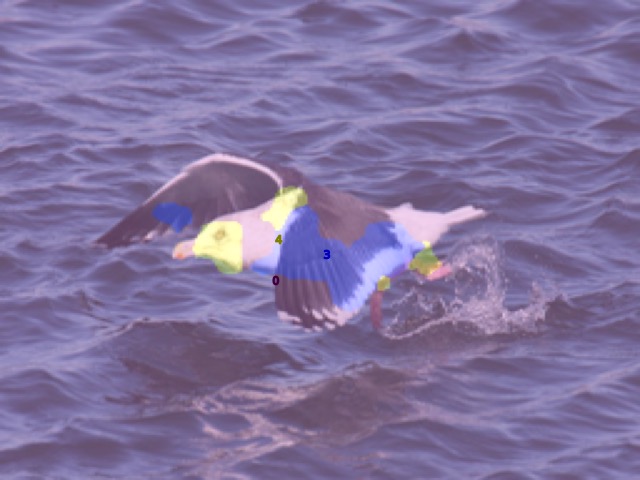}  &
 \includegraphics[width=1.6 cm, height=1.6 cm]{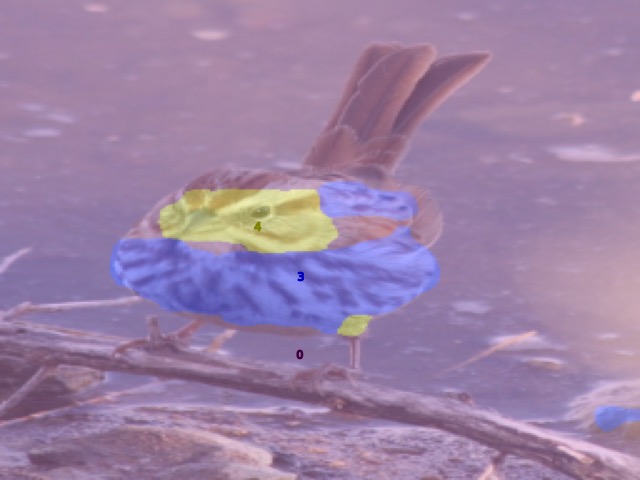}  &
 \includegraphics[width=1.6 cm, height=1.6 cm]{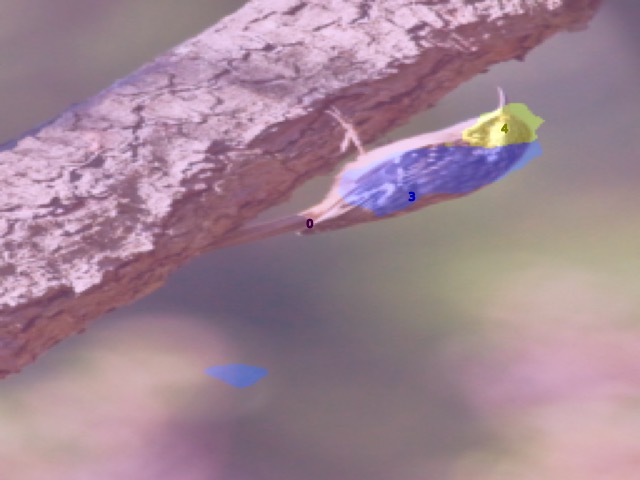}  \\

\rotatebox{90}{PDiscoNet} &
  \includegraphics[width=1.6 cm, height=1.6 cm]{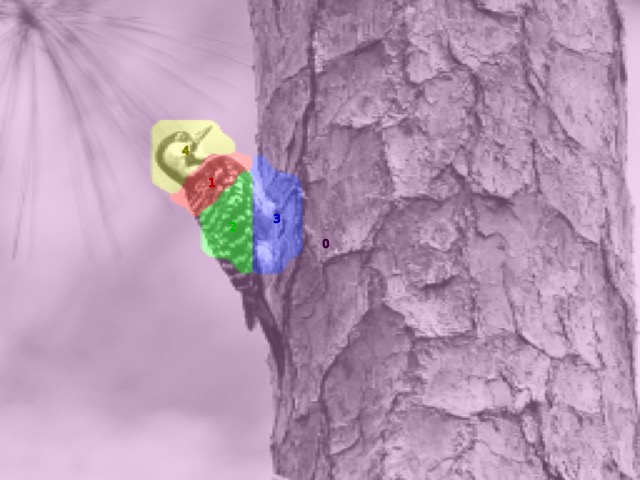} &
 \includegraphics[width=1.6 cm, height=1.6 cm]{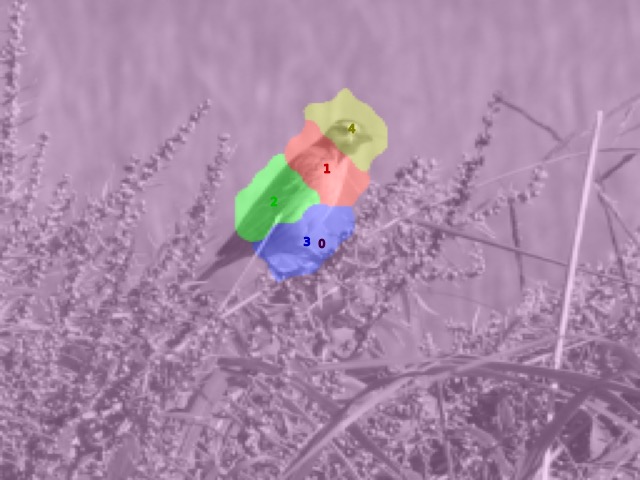}  &
 \includegraphics[width=1.6 cm, height=1.6 cm]{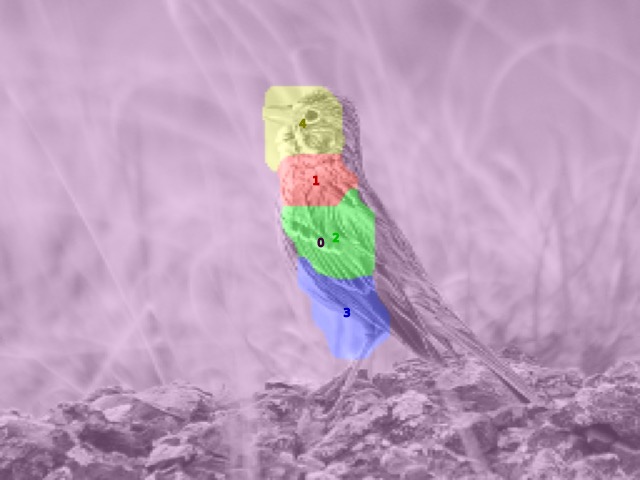}  &
 \includegraphics[width=1.6 cm, height=1.6 cm]{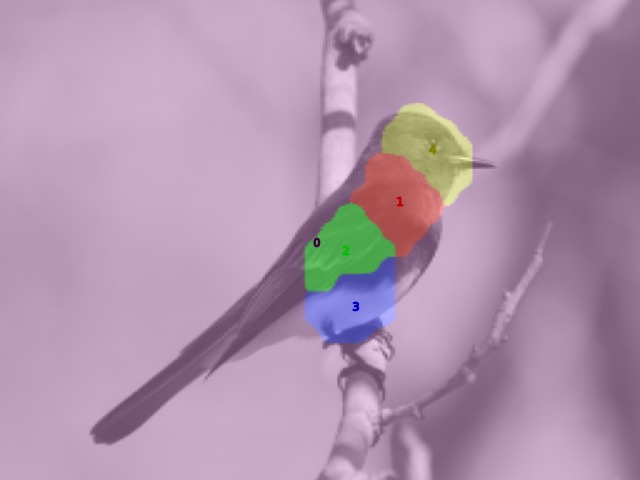}  &
 \includegraphics[width=1.6 cm, height=1.6 cm]{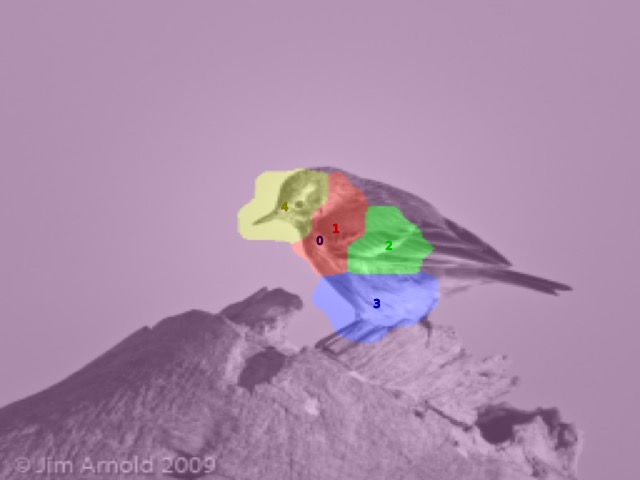}  &
 \includegraphics[width=1.6 cm, height=1.6 cm]{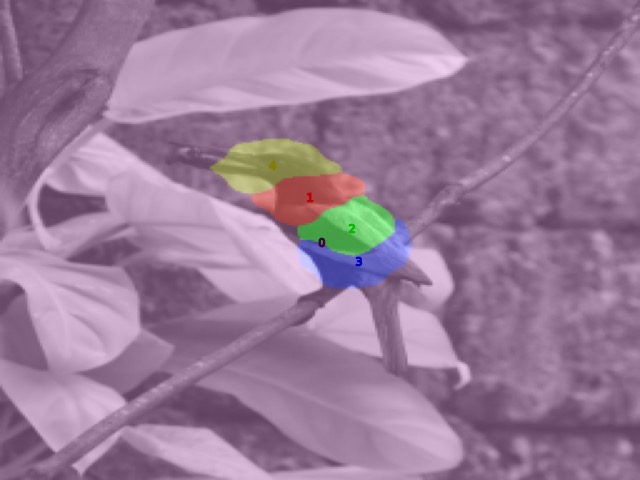}  &
 \includegraphics[width=1.6 cm, height=1.6 cm]{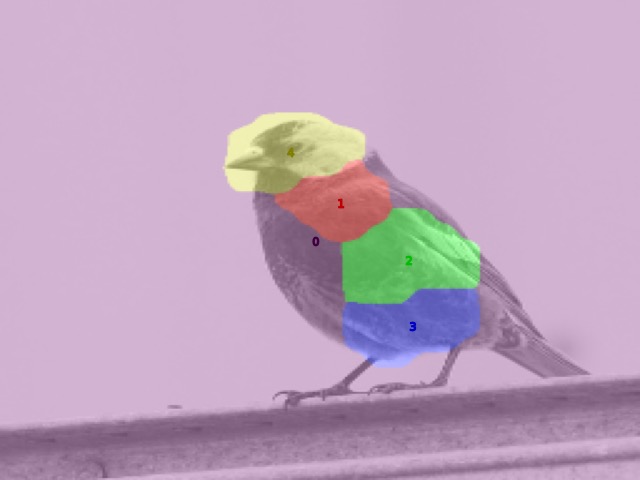}  &
 \includegraphics[width=1.6 cm, height=1.6 cm]{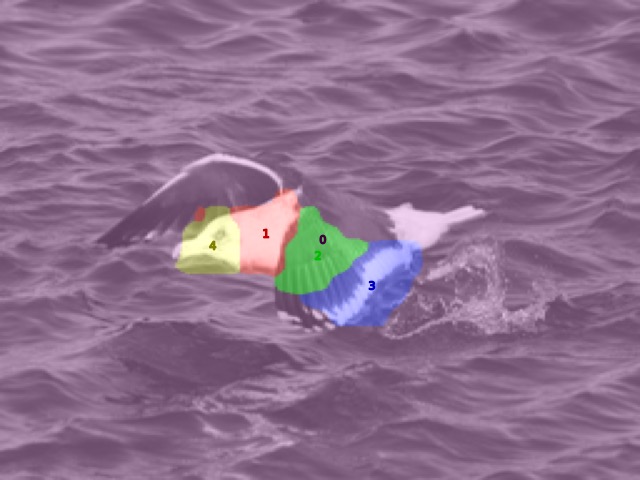}  &
 \includegraphics[width=1.6 cm, height=1.6 cm]{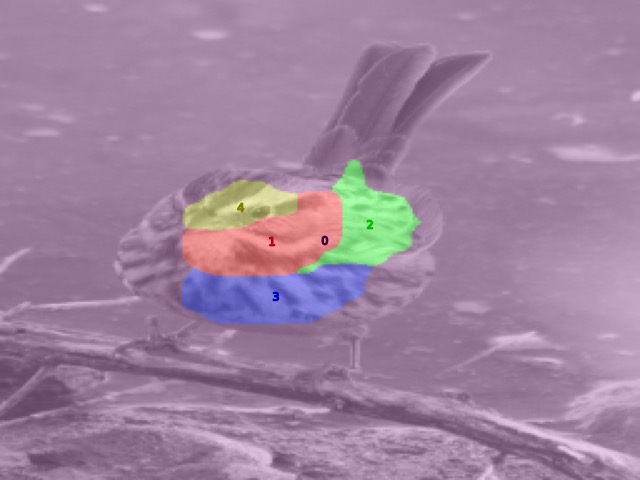}  &
 \includegraphics[width=1.6 cm, height=1.6 cm]{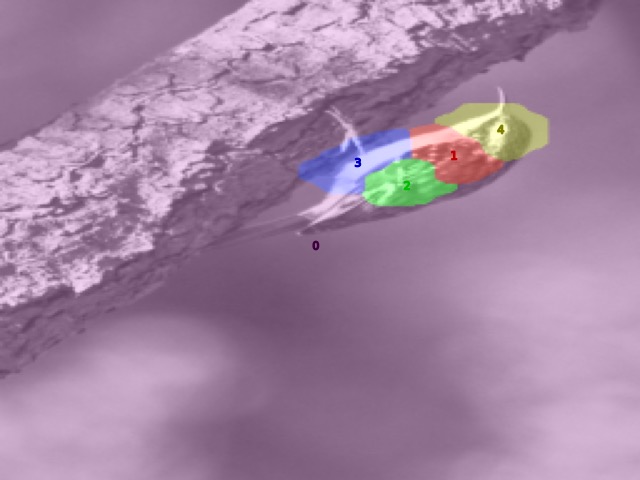}  \\
  
 \end{tabular}
  \caption{Discovered part segmentation with $K=4$ on CUB for \cite{amir_deep_2022_dinovit} (top), \cite{huang_interpretable_2020} (middle) and  our method (bottom). }
  \label{fig:qual_cub2}
\end{figure*}

\begin{figure*}
\centering
\setlength\tabcolsep{1.5pt} 
 \begin{tabular}{lcccccccccc}
 \rotatebox{90}{Original} &
 \includegraphics[width=1.6 cm, height=1.6 cm]{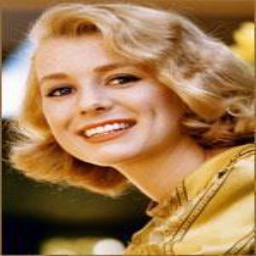} &
 \includegraphics[width=1.6 cm, height=1.6 cm]{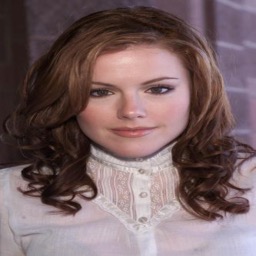}  &
 \includegraphics[width=1.6 cm, height=1.6 cm]{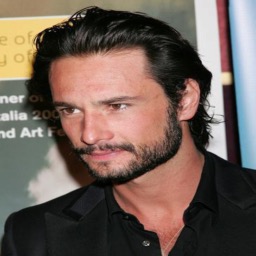}  &
 \includegraphics[width=1.6 cm, height=1.6 cm]{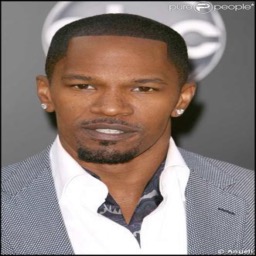}  &
 \includegraphics[width=1.6 cm, height=1.6 cm]{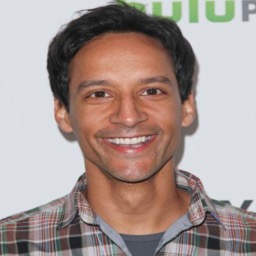}  &
 \includegraphics[width=1.6 cm, height=1.6 cm]{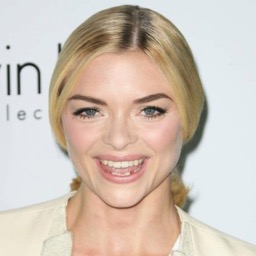}  &
 \includegraphics[width=1.6 cm, height=1.6 cm]{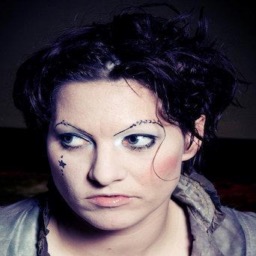}  &
 \includegraphics[width=1.6 cm, height=1.6 cm]{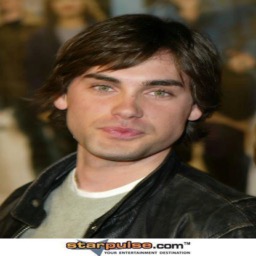}  &
 \includegraphics[width=1.6 cm, height=1.6 cm]{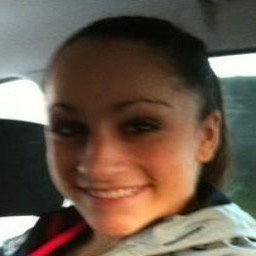}  &
 \includegraphics[width=1.6 cm, height=1.6 cm]{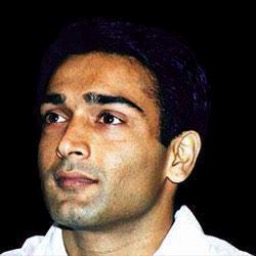}  \\

\rotatebox{90}{Dino~\cite{amir_deep_2022_dinovit}} &
    \includegraphics[width=1.6 cm, height=1.6 cm]{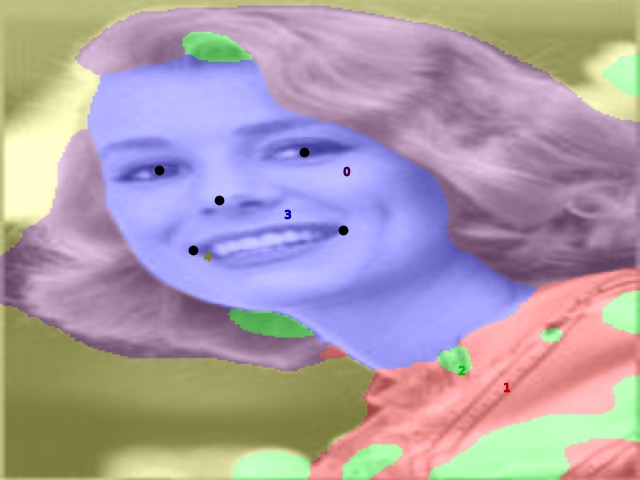} &
 \includegraphics[width=1.6 cm, height=1.6 cm]{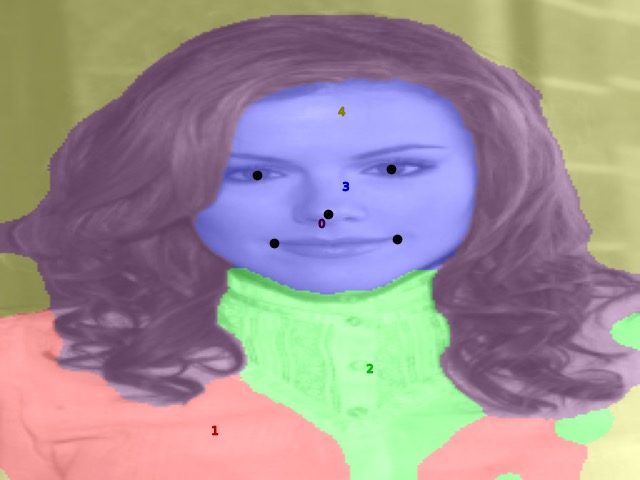}  &
 \includegraphics[width=1.6 cm, height=1.6 cm]{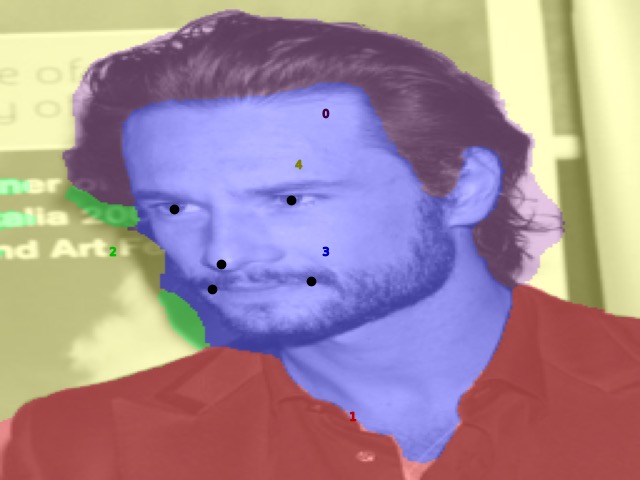}  &
 \includegraphics[width=1.6 cm, height=1.6 cm]{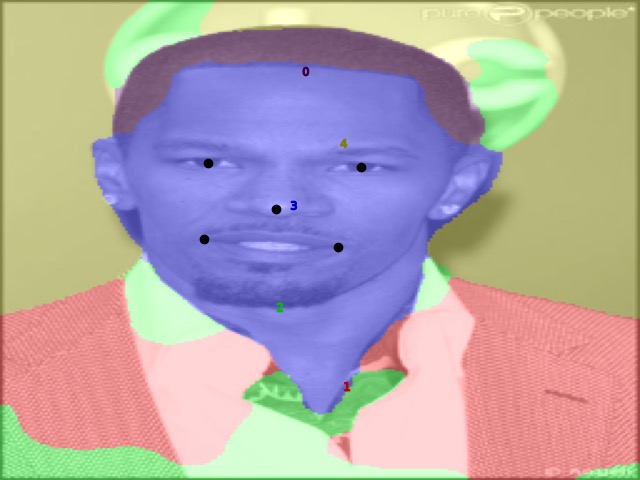}  &
 \includegraphics[width=1.6 cm, height=1.6 cm]{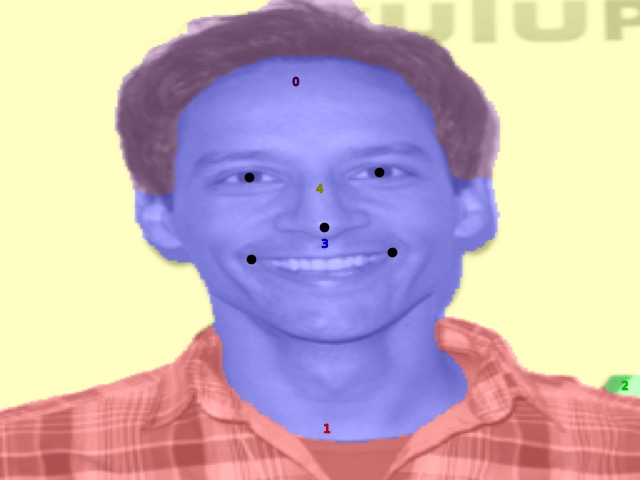}  &
 \includegraphics[width=1.6 cm, height=1.6 cm]{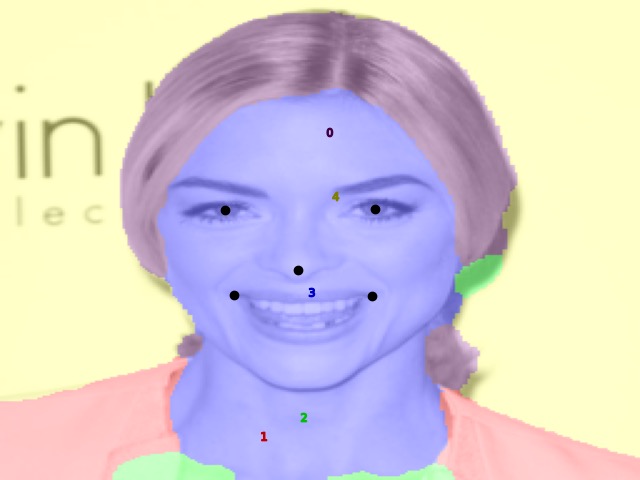}  &
 \includegraphics[width=1.6 cm, height=1.6 cm]{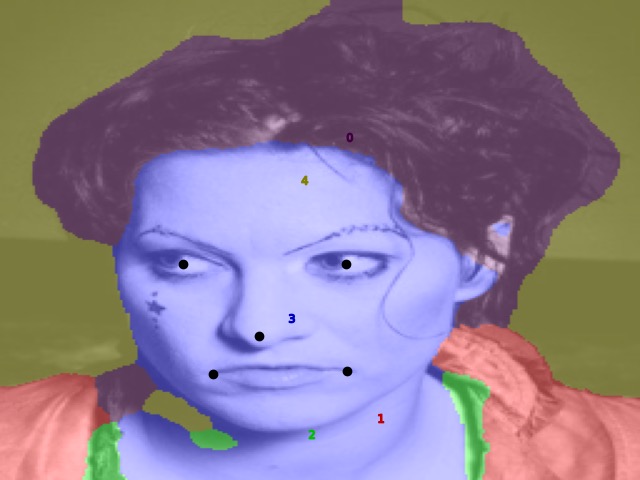}  &
 \includegraphics[width=1.6 cm, height=1.6 cm]{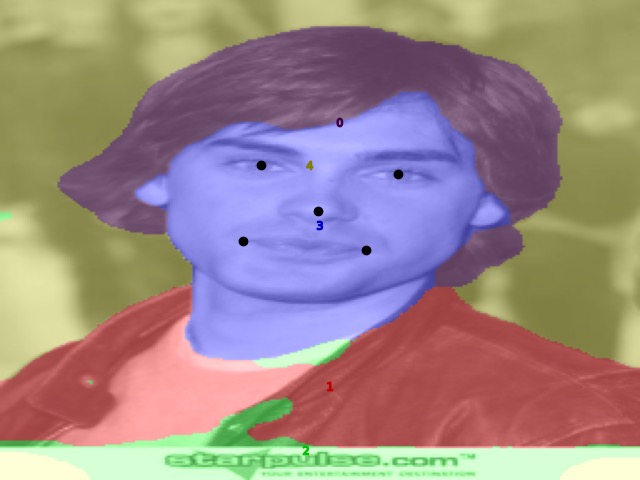}  &
 \includegraphics[width=1.6 cm, height=1.6 cm]{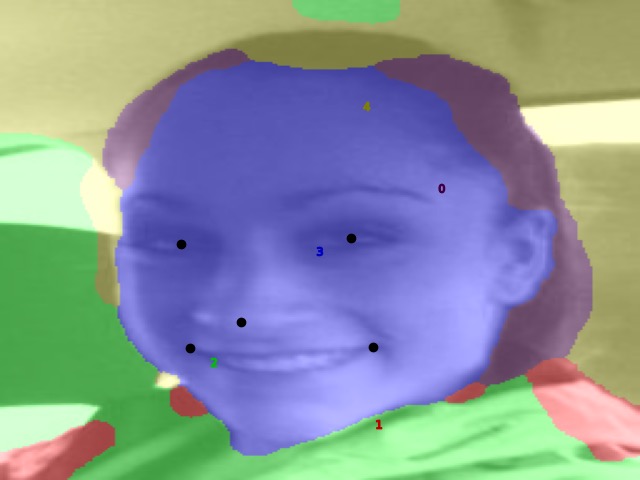}  &
 \includegraphics[width=1.6 cm, height=1.6 cm]{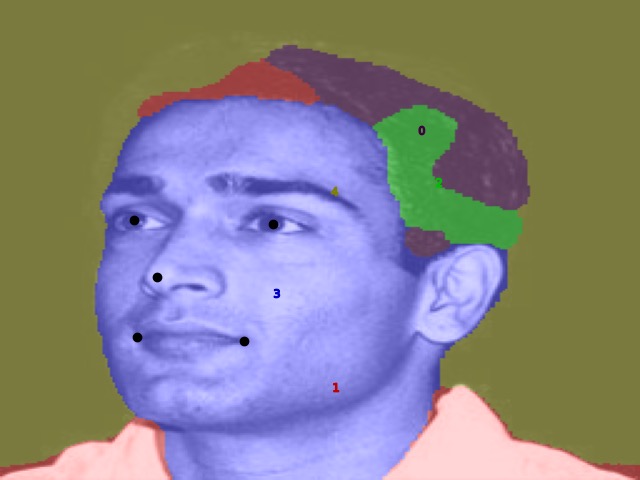}  \\

\rotatebox{90}{Huang~\cite{huang_interpretable_2020}} &
  \includegraphics[width=1.6 cm, height=1.6 cm]{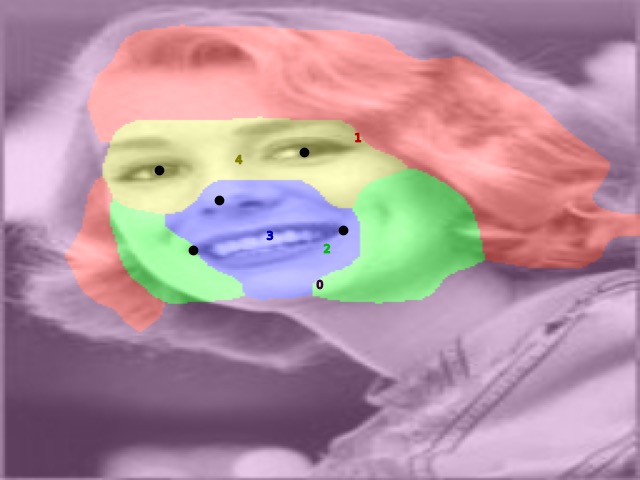} &
 \includegraphics[width=1.6 cm, height=1.6 cm]{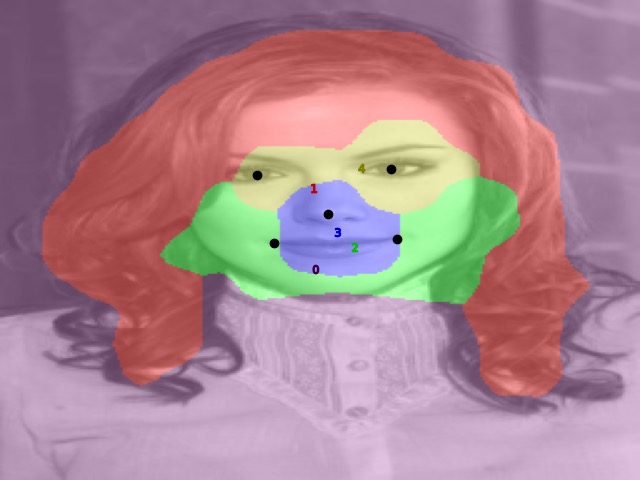}  &
 \includegraphics[width=1.6 cm, height=1.6 cm]{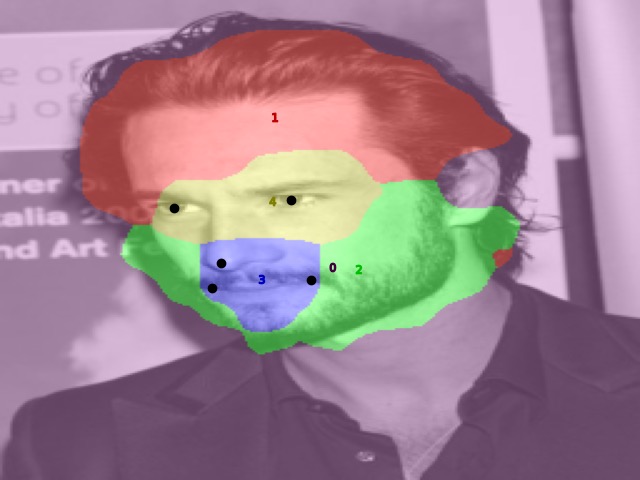}  &
 \includegraphics[width=1.6 cm, height=1.6 cm]{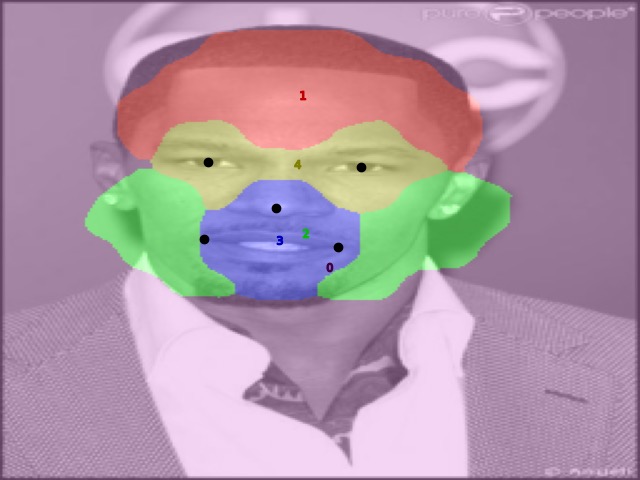}  &
 \includegraphics[width=1.6 cm, height=1.6 cm]{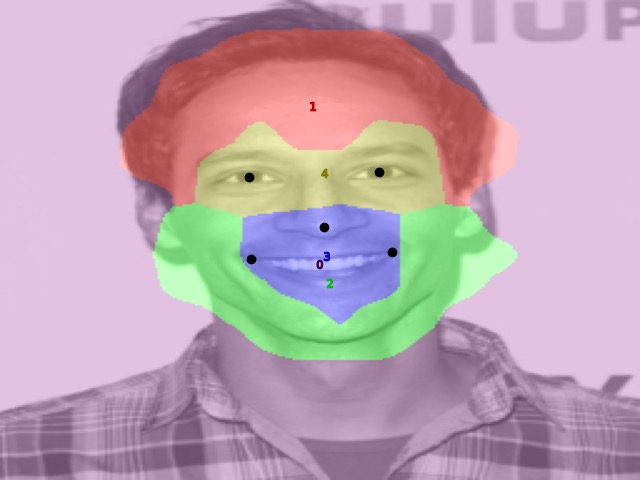}  &
 \includegraphics[width=1.6 cm, height=1.6 cm]{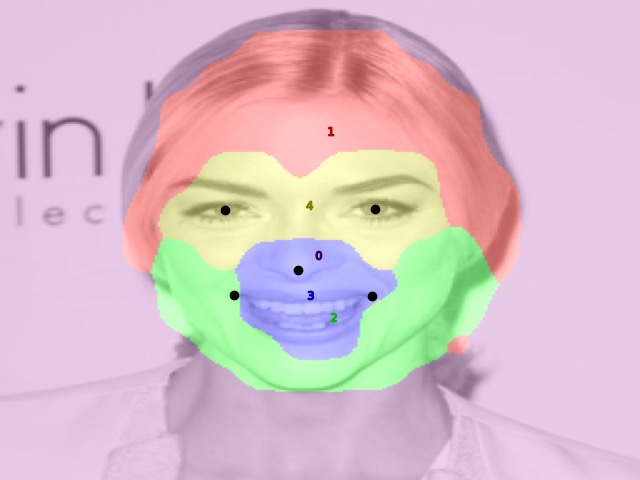}  &
 \includegraphics[width=1.6 cm, height=1.6 cm]{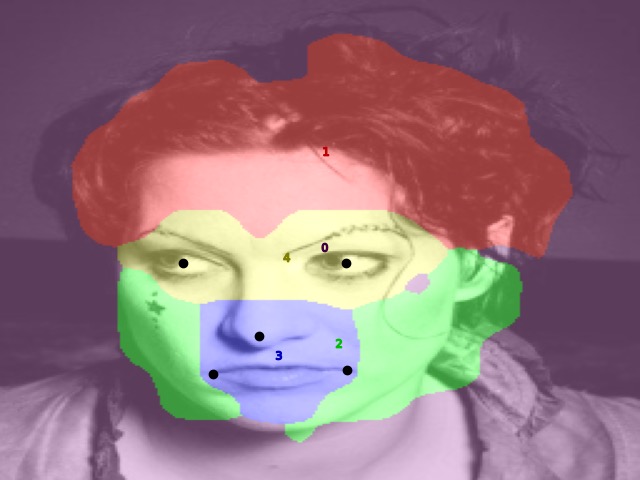}  &
 \includegraphics[width=1.6 cm, height=1.6 cm]{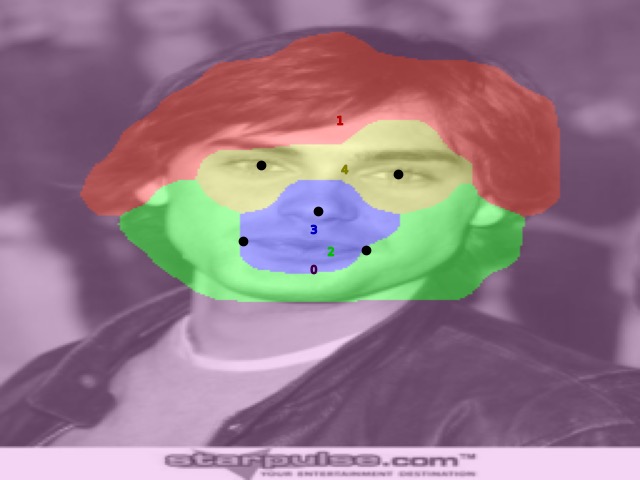}  &
 \includegraphics[width=1.6 cm, height=1.6 cm]{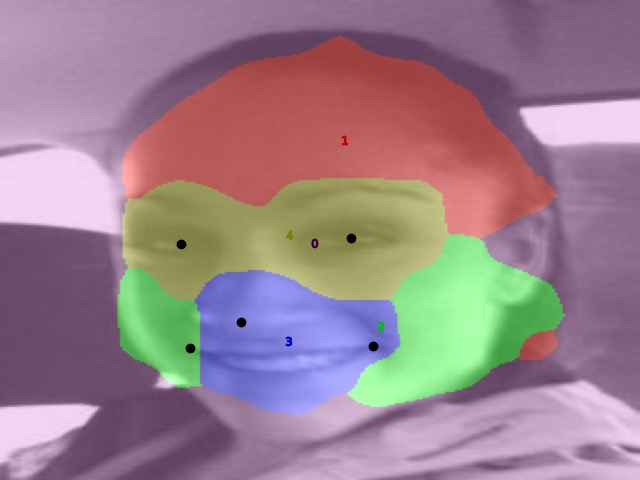}  &
 \includegraphics[width=1.6 cm, height=1.6 cm]{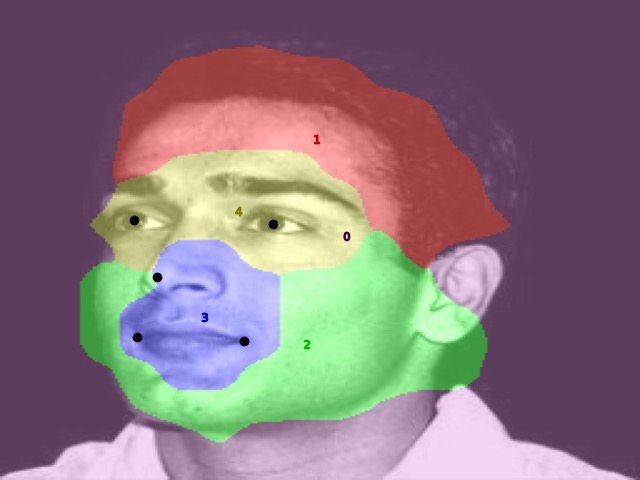}  \\

\rotatebox{90}{PDiscoNet} &
  \includegraphics[width=1.6 cm, height=1.6 cm]{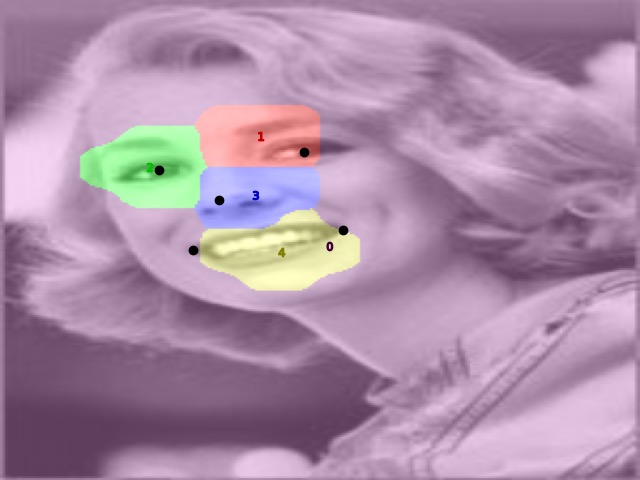} &
 \includegraphics[width=1.6 cm, height=1.6 cm]{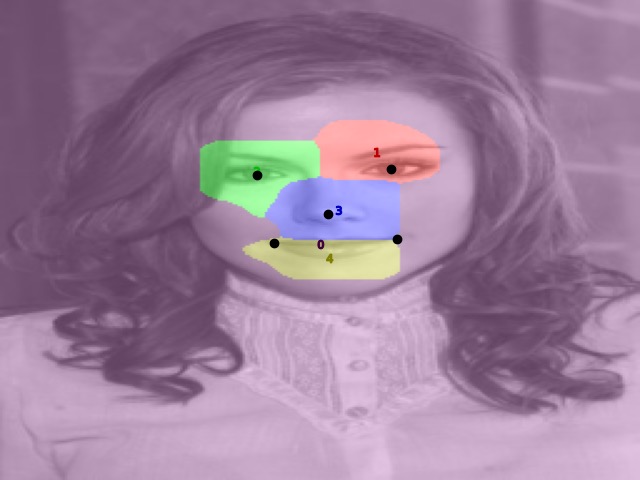}  &
 \includegraphics[width=1.6 cm, height=1.6 cm]{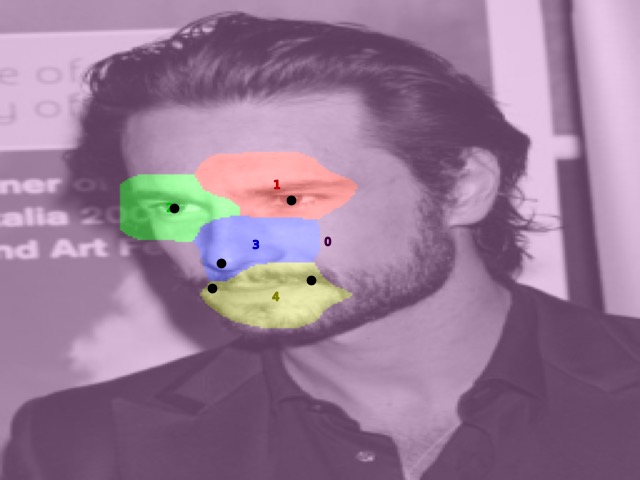}  &
 \includegraphics[width=1.6 cm, height=1.6 cm]{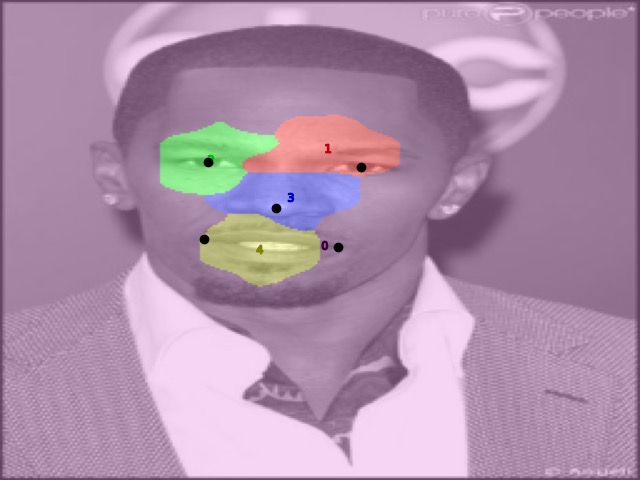}  &
 \includegraphics[width=1.6 cm, height=1.6 cm]{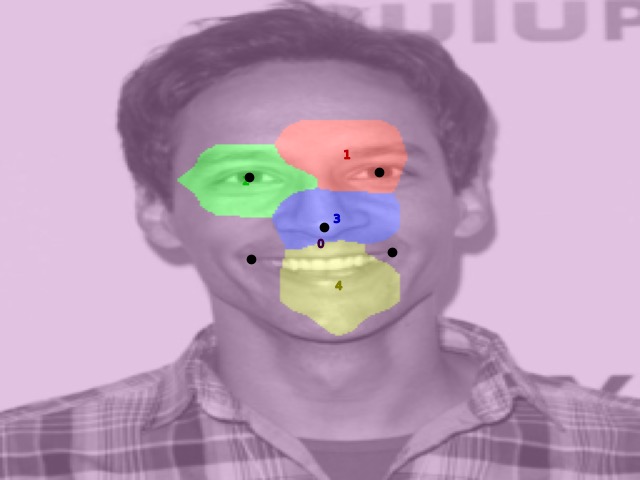}  &
 \includegraphics[width=1.6 cm, height=1.6 cm]{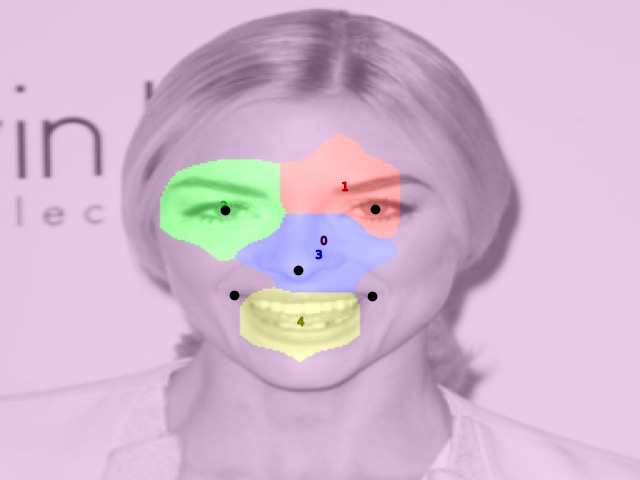}  &
 \includegraphics[width=1.6 cm, height=1.6 cm]{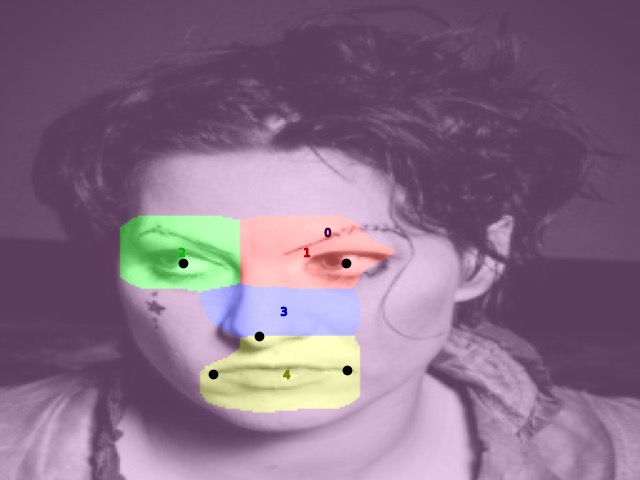}  &
 \includegraphics[width=1.6 cm, height=1.6 cm]{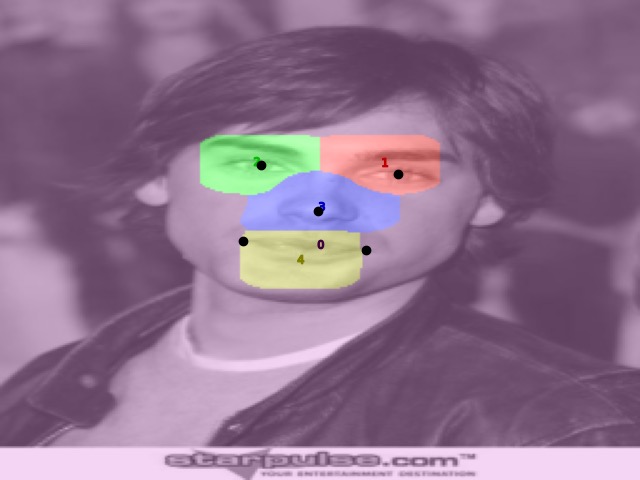}  &
 \includegraphics[width=1.6 cm, height=1.6 cm]{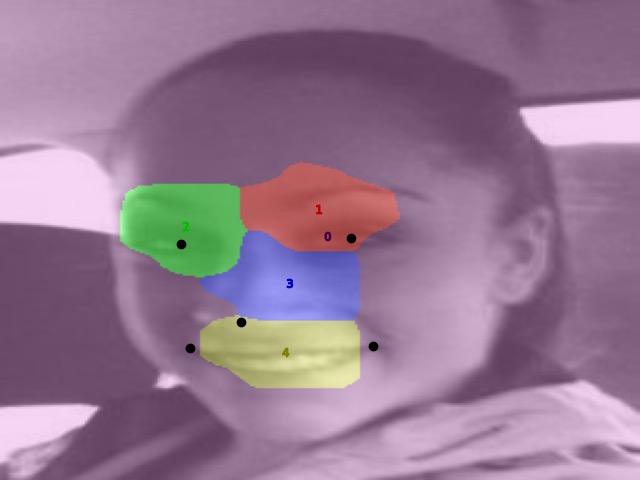}  &
 \includegraphics[width=1.6 cm, height=1.6 cm]{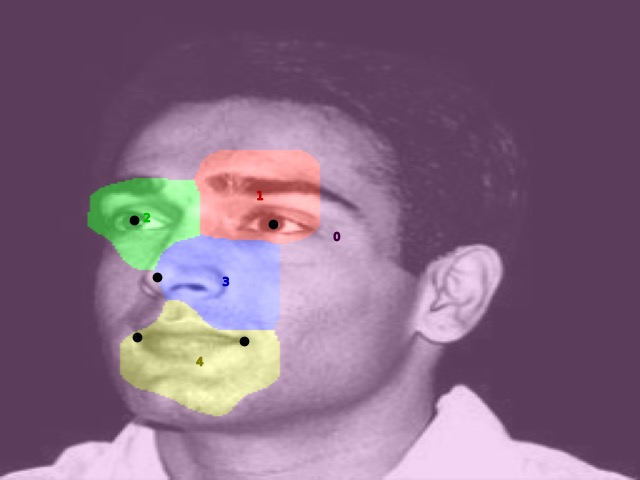}  \\
  
 \end{tabular}
  \caption{Discovered part segmentation with $K=4$ on CelebA for \cite{amir_deep_2022_dinovit} (top), \cite{huang_interpretable_2020} (middle) and  our method (bottom).  The ground truth facial landmarks appear as black dots.}
  \label{fig:qual_celeb2}
\end{figure*}

\begin{figure*}
\centering
\setlength\tabcolsep{1.5pt} 
 \begin{tabular}{lcccccccccc}
  \rotatebox{90}{Original} &
 \includegraphics[width=1.6 cm, height=1.6 cm]{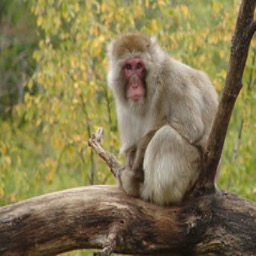} &
 \includegraphics[width=1.6 cm, height=1.6 cm]{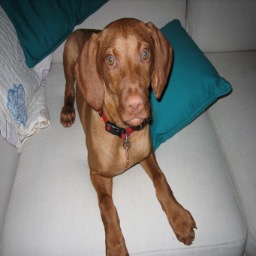}  &
 \includegraphics[width=1.6 cm, height=1.6 cm]{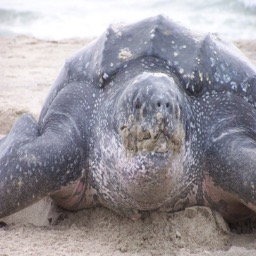}  &
 \includegraphics[width=1.6 cm, height=1.6 cm]{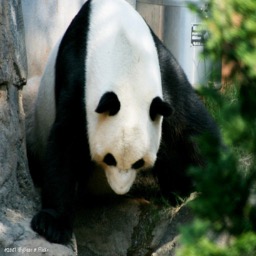}  &
 \includegraphics[width=1.6 cm, height=1.6 cm]{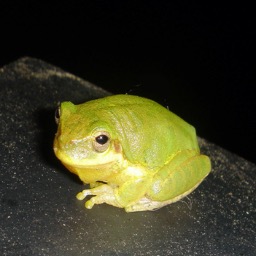}  &
 \includegraphics[width=1.6 cm, height=1.6 cm]{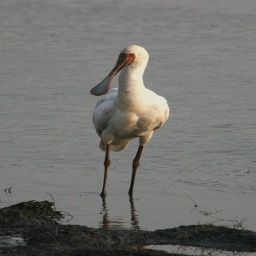}  &
 \includegraphics[width=1.6 cm, height=1.6 cm]{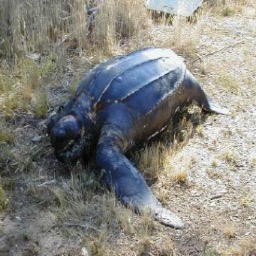}  &
 \includegraphics[width=1.6 cm, height=1.6 cm]{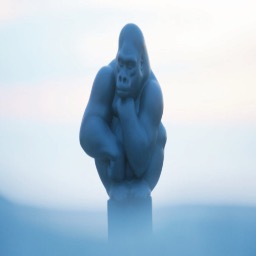}  &
 \includegraphics[width=1.6 cm, height=1.6 cm]{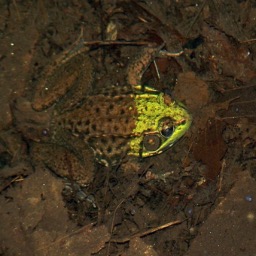}  &
 \includegraphics[width=1.6 cm, height=1.6 cm]{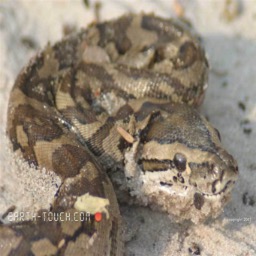}  \\

\rotatebox{90}{Dino~\cite{amir_deep_2022_dinovit}} &
     \includegraphics[width=1.6 cm, height=1.6 cm]{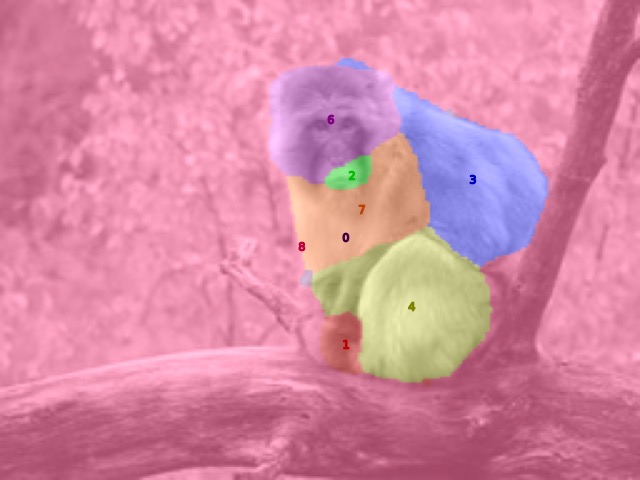} &
 \includegraphics[width=1.6 cm, height=1.6 cm]{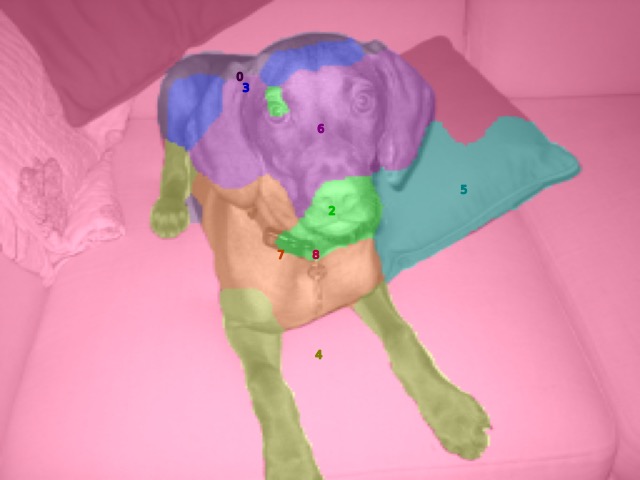}  &
 \includegraphics[width=1.6 cm, height=1.6 cm]{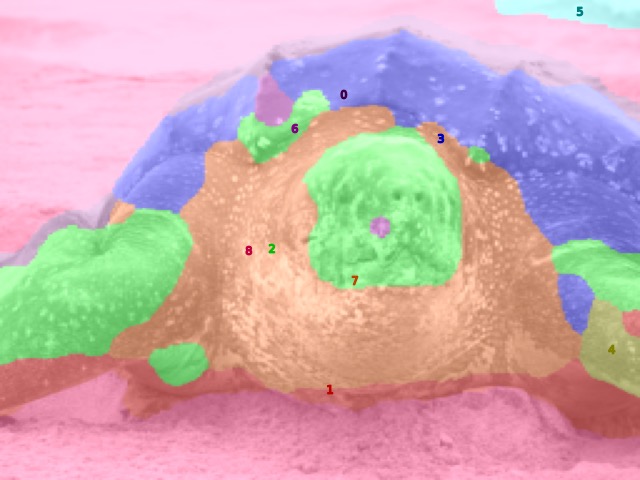}  &
 \includegraphics[width=1.6 cm, height=1.6 cm]{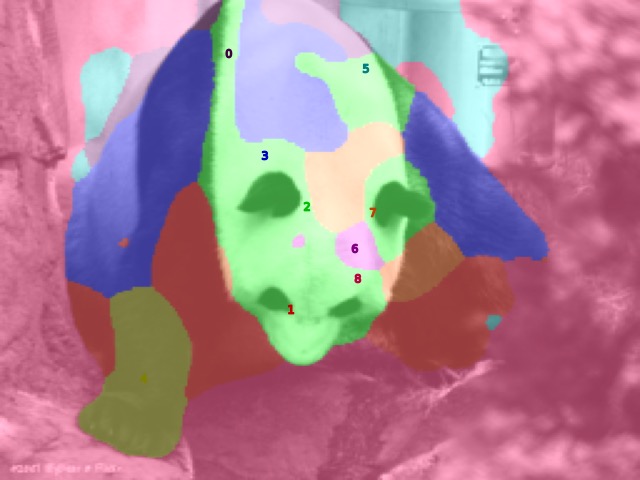}  &
 \includegraphics[width=1.6 cm, height=1.6 cm]{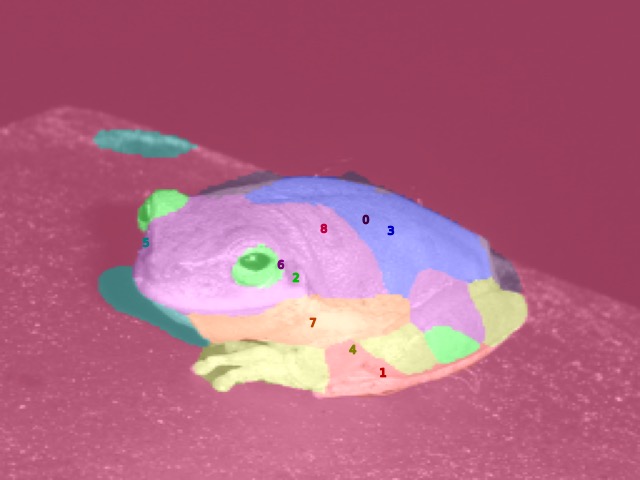}  &
 \includegraphics[width=1.6 cm, height=1.6 cm]{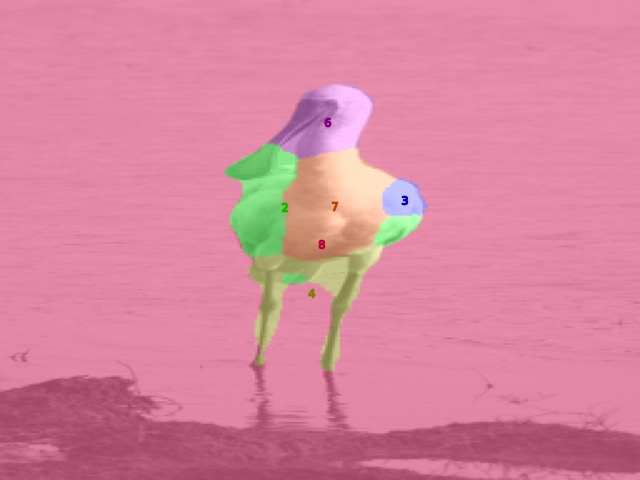}  &
 \includegraphics[width=1.6 cm, height=1.6 cm]{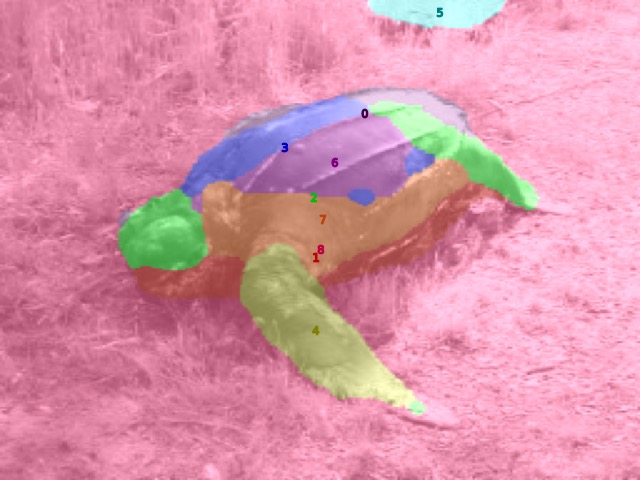}  &
 \includegraphics[width=1.6 cm, height=1.6 cm]{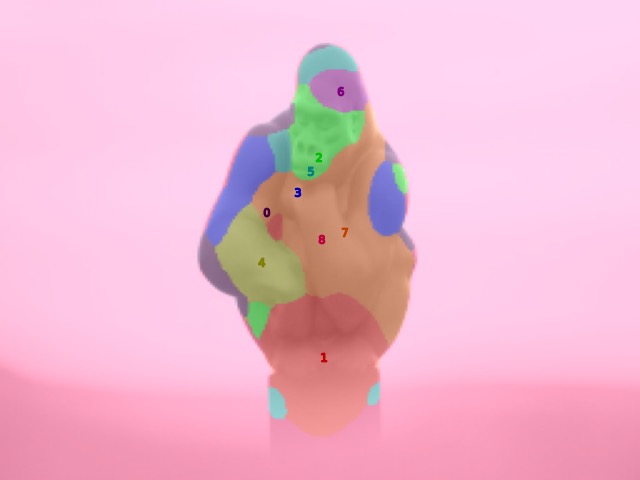}  &
 \includegraphics[width=1.6 cm, height=1.6 cm]{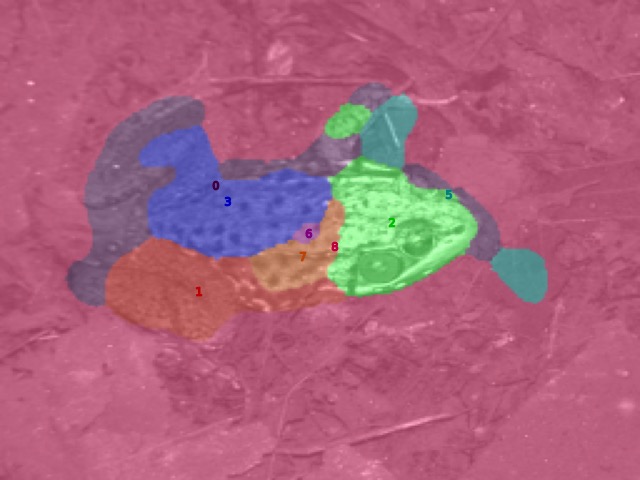}  &
 \includegraphics[width=1.6 cm, height=1.6 cm]{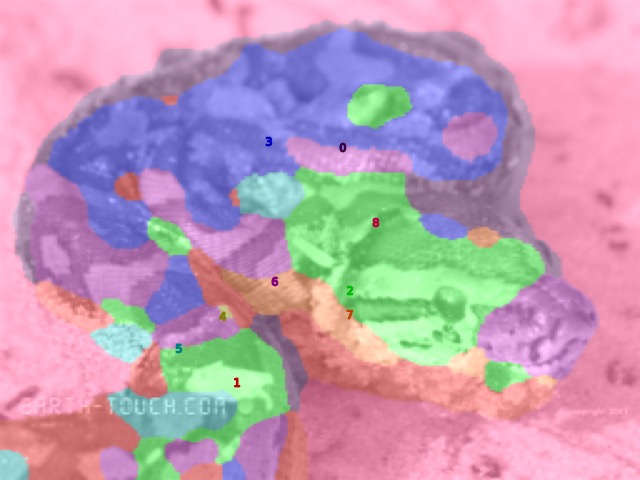}  \\

\rotatebox{90}{Huang~\cite{huang_interpretable_2020}} &
   \includegraphics[width=1.6 cm, height=1.6 cm]{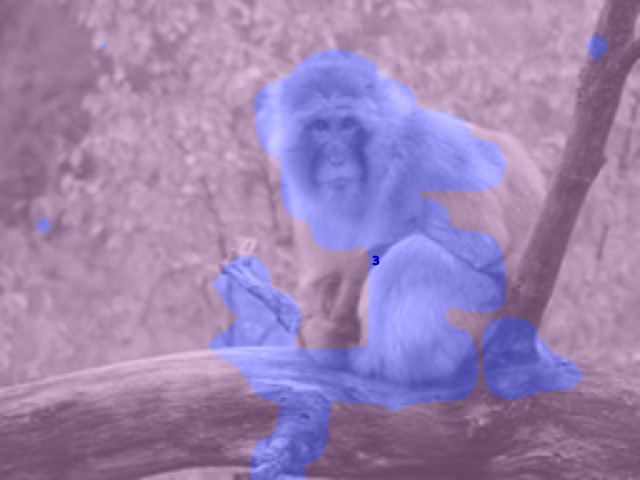} &
 \includegraphics[width=1.6 cm, height=1.6 cm]{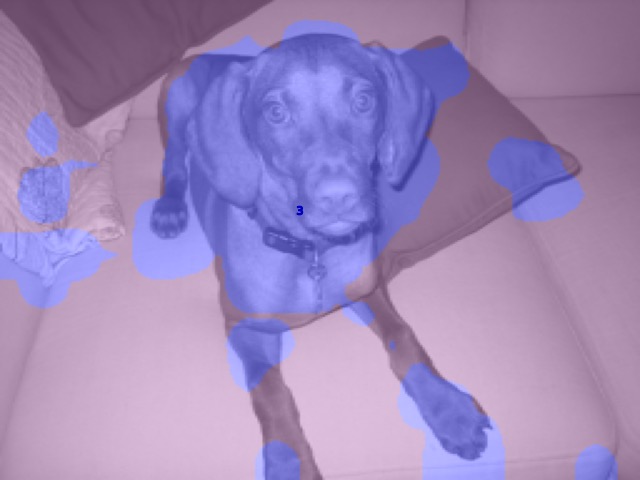}  &
 \includegraphics[width=1.6 cm, height=1.6 cm]{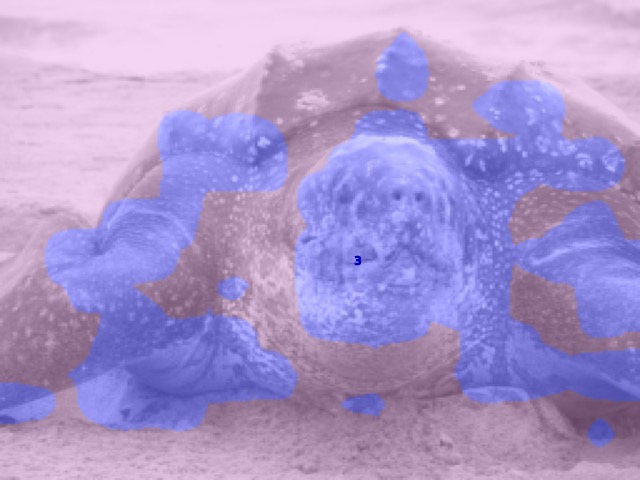}  &
 \includegraphics[width=1.6 cm, height=1.6 cm]{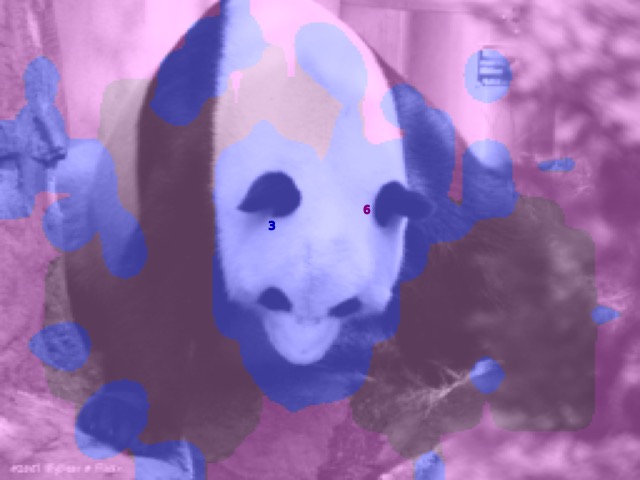}  &
 \includegraphics[width=1.6 cm, height=1.6 cm]{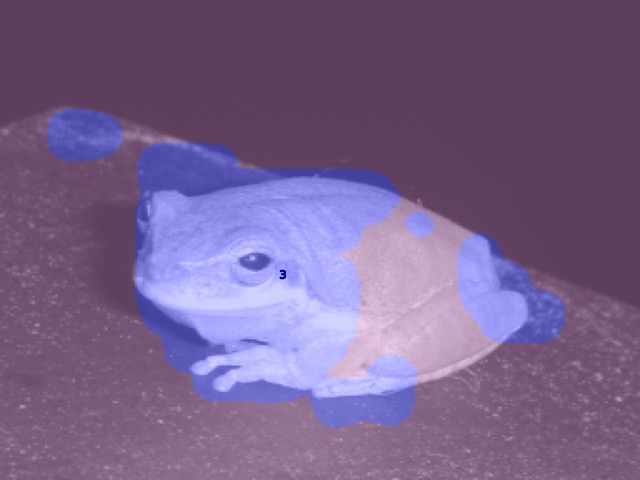}  &
 \includegraphics[width=1.6 cm, height=1.6 cm]{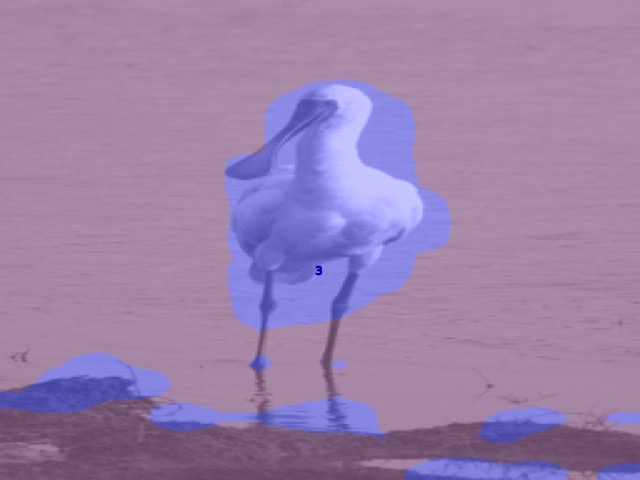}  &
 \includegraphics[width=1.6 cm, height=1.6 cm]{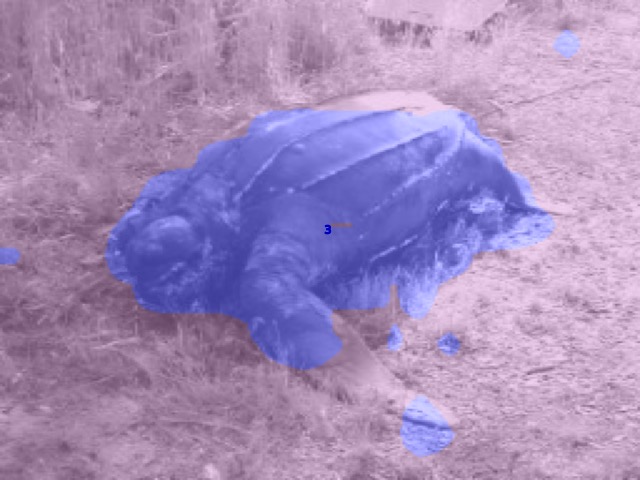}  &
 \includegraphics[width=1.6 cm, height=1.6 cm]{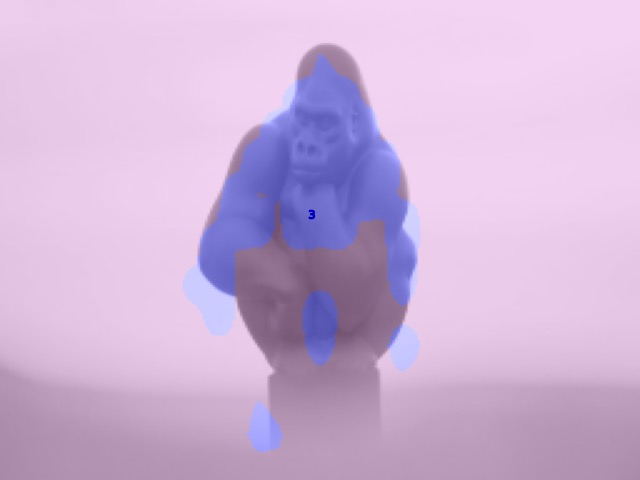}  &
 \includegraphics[width=1.6 cm, height=1.6 cm]{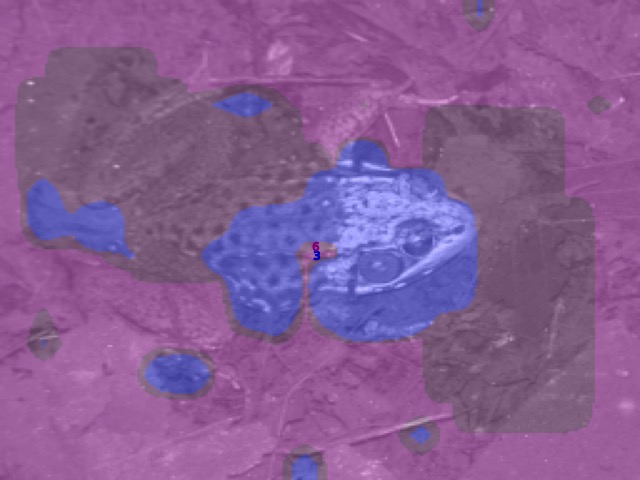}  &
 \includegraphics[width=1.6 cm, height=1.6 cm]{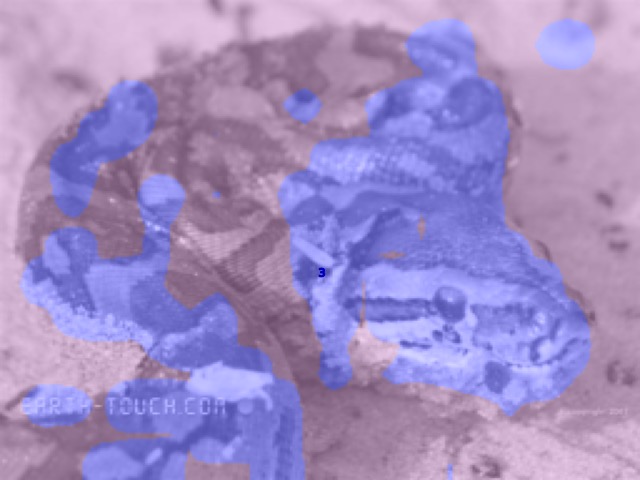}  \\

 \rotatebox{90}{PDiscoNet} &
  \includegraphics[width=1.6 cm, height=1.6 cm]{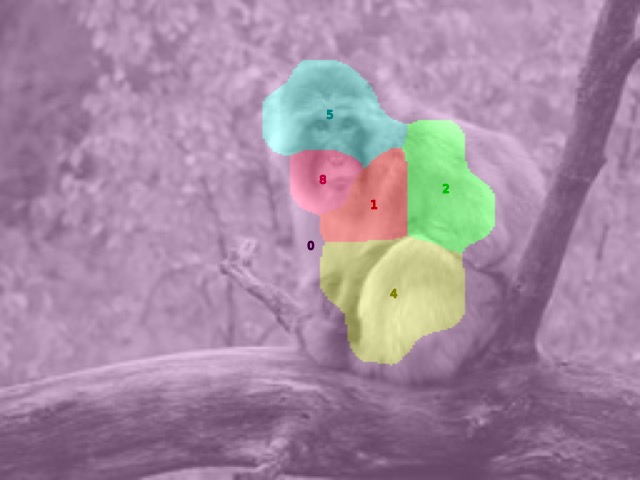} &
 \includegraphics[width=1.6 cm, height=1.6 cm]{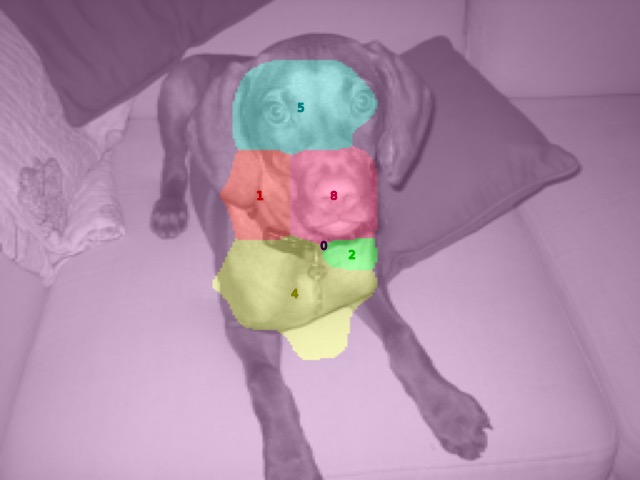}  &
 \includegraphics[width=1.6 cm, height=1.6 cm]{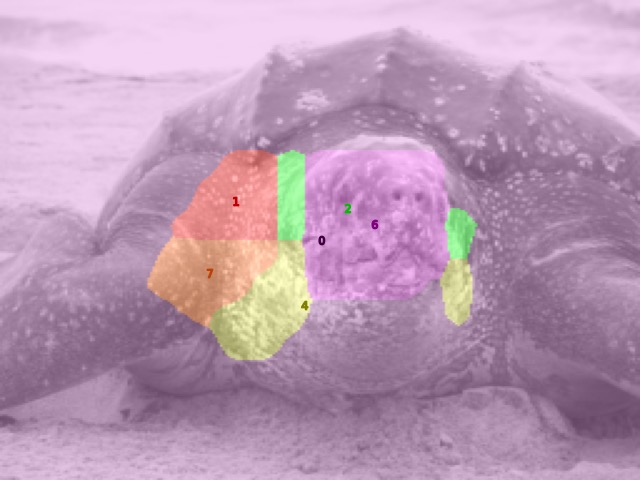}  &
 \includegraphics[width=1.6 cm, height=1.6 cm]{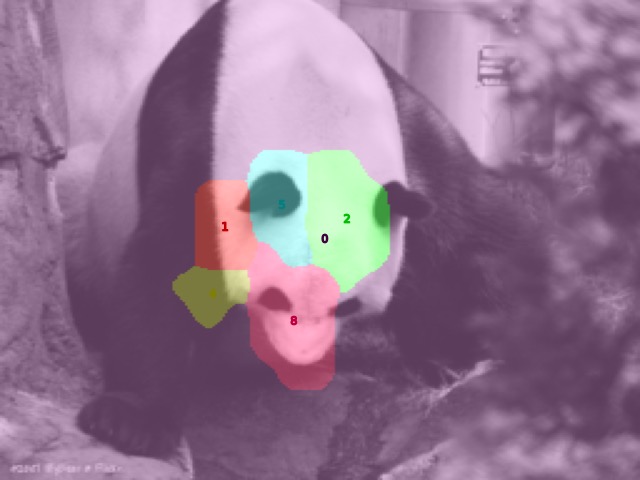}  &
 \includegraphics[width=1.6 cm, height=1.6 cm]{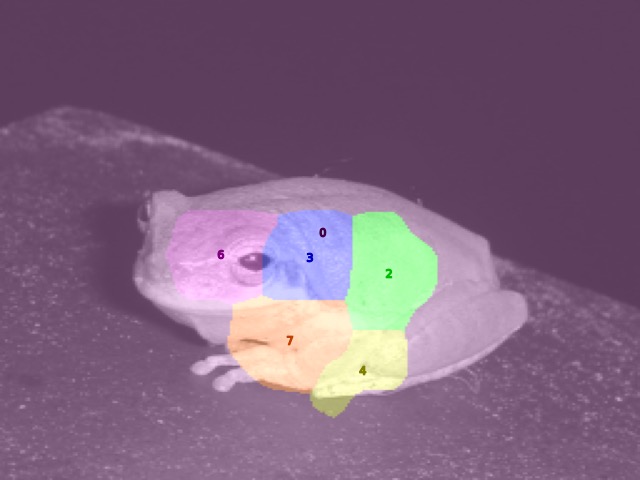}  &
 \includegraphics[width=1.6 cm, height=1.6 cm]{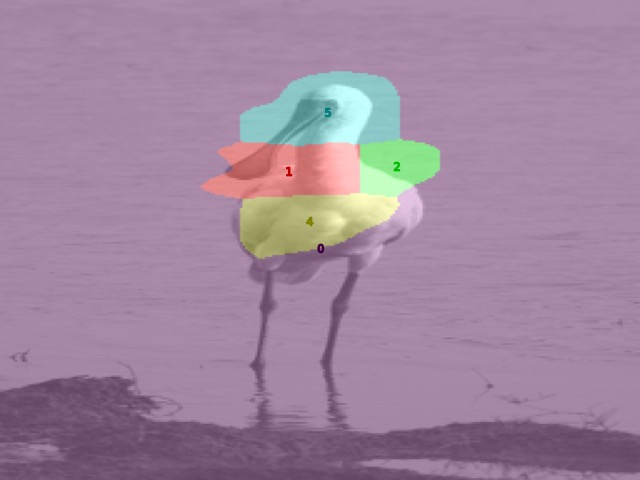}  &
 \includegraphics[width=1.6 cm, height=1.6 cm]{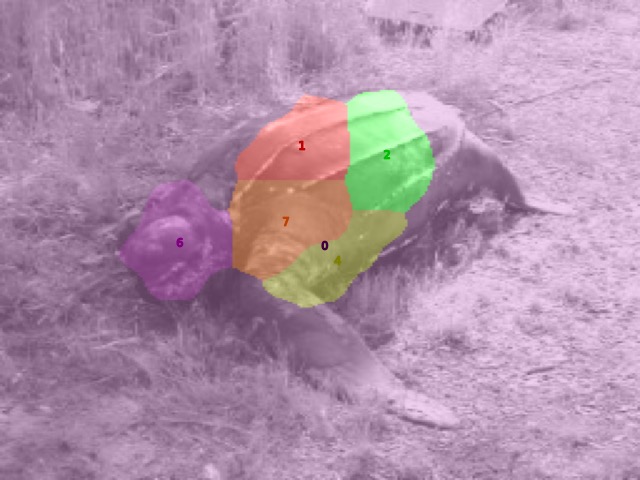}  &
 \includegraphics[width=1.6 cm, height=1.6 cm]{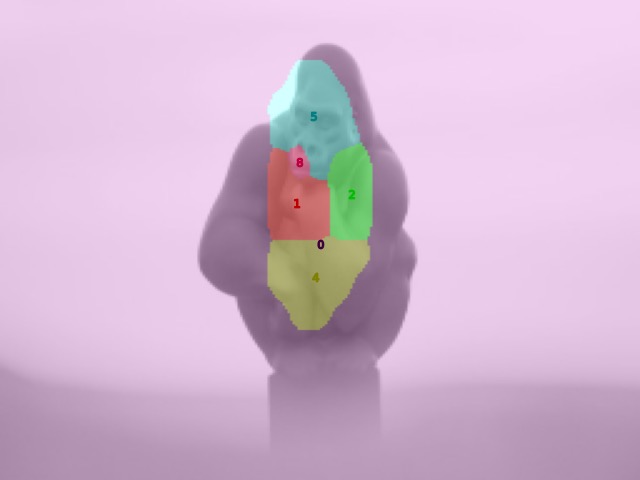}  &
 \includegraphics[width=1.6 cm, height=1.6 cm]{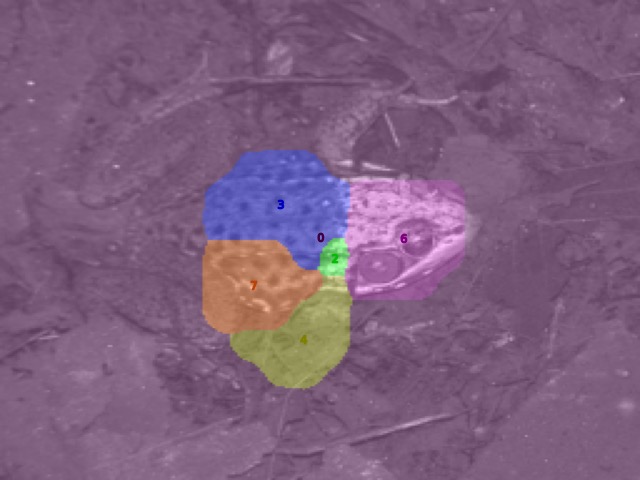}  &
 \includegraphics[width=1.6 cm, height=1.6 cm]{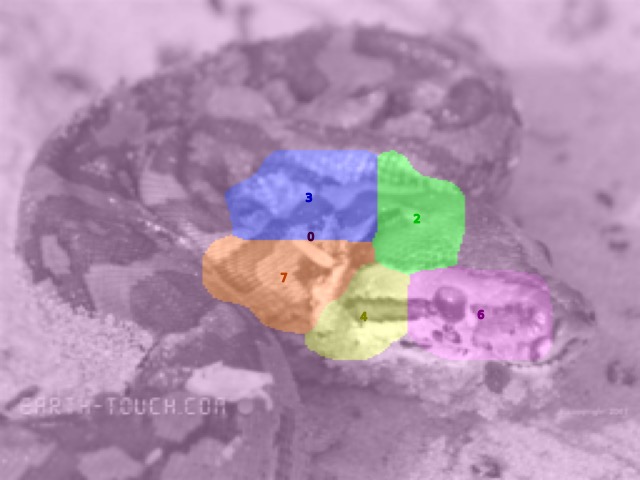}  \\

 \end{tabular}
  \caption{Discovered part segmentation with $K=8$ on PartImageNet for \cite{amir_deep_2022_dinovit} (top), \cite{huang_interpretable_2020} (middle) and  our method (bottom).  }
  \label{fig:qual_pim2}
\end{figure*}

\begin{figure*}[t]
    \centering
    \includegraphics[width=\textwidth]{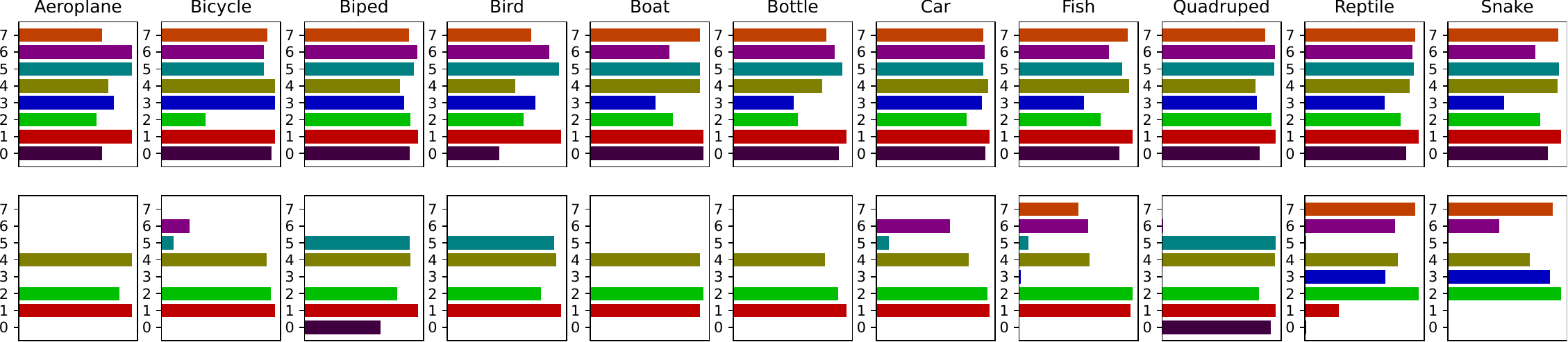}
    \caption{Histograms of part presence per PartImageNet supercategory for Dino ViT~\cite{amir_deep_2022_dinovit} (top) and our method (bottom), with $K=8$. Same color code as the figure above.}
    \label{fig:parts_cat}
\end{figure*}

In  Figs.~\ref{fig:qual_cub2} to~\ref{fig:qual_celeb2} we show the assignment maps for the three methods, with an attention threshold of 0.1 for~\cite{huang_interpretable_2020}, across ten images for each dataset in order to explore the semantic consistency of the discovered parts across a diverse set of examples.
As can be seen in Fig.~\ref{fig:qual_celeb2}, \cite{huang_interpretable_2020} tends to find symmetric part assignment maps, while our method finds independent parts for the areas around each eye. This figure also explains why \cite{amir_deep_2022_dinovit} fails in the clustering metrics: the whole face tends to be assigned to a single part, making all the facial landmarks indistinguishable from each other. This phenomenon showcases that, although the self-supervised approach of~\cite{amir_deep_2022_dinovit} provides remarkable results in terms of semantics and boundary adherence, it may also miss the relevant partitions due to not making use of the fine-grained recognition signal.
This drawback of the Dino ViT approach is also visible in the CUB results in Fig.~\ref{fig:qual_cub2}, where in some examples some parts either mix with background elements (first column) or are completely missed (last column), while our method is consistent across all samples.
The method by Huang and Li~\cite{huang_interpretable_2020} also displays a problem with mixing in background parts in CUB and, even more markedly, in PartImageNet. Fig.~\ref{fig:qual_cub} shows that a majority of the available parts tend to be used on background areas that ultimately receive a low attention weight, leading to only two or three parts being used even in the case of $K=16$. In Fig.~\ref{fig:qual_pim} we can see that~\cite{huang_interpretable_2020} assigns one single part to foreground objects with $K=8$ and fails to assign any parts to foreground objects for $K=25$, which explains the low part discovery and classification scores in Table~\ref{tab:quant}.
Both our method and~\cite{amir_deep_2022_dinovit} are generally able to identify the object of interest in PartImagenet (Figs.~\ref{fig:qual_pim} and~\ref{fig:qual_pim2}), with~\cite{amir_deep_2022_dinovit} often providing better boundary adherence, while our method tends to provide better semantic consistency. For instance, notice how our method uses the same part (in cyan) for the head of mammals and birds, while another one (in purple) is used for the head of reptiles and amphibians.

This better semantic consistency is further reinforced by Fig.~\ref{fig:parts_cat}, where we show the histograms of part presence per PartImageNet supercategory for Dino ViT~\cite{amir_deep_2022_dinovit} and our method with $K=8$. We omit the results on~\cite{huang_interpretable_2020} because only one part tended to be active in high attention areas (see Fig.~\ref{fig:qual_pim2}).
This figure confirms the notion that PDiscoNet discovers parts with strong semantic consitency. We can see that similar supercategories (such as \emph{Aeroplane} and \emph{Boat}, or \emph{Biped} and \emph{Quadruped}) tend to share the same parts, and parts tend to specialize on only a subset of supercategories. For instance, we note that the cyan part (number 5) is indeed mostly present in \emph{Biped}, \emph{Quadruped} and \emph{Bird}, while the orange part (number 7), is only present in \emph{Fish}, \emph{Reptile} and \emph{Snake}. On the other hand, all parts are almost equally shared by all supercategories in the case of Dino ViT, indicating that parts are less semantically consistent across the dataset and acquire multiple semantic interpretations.

The quantitative and qualitative results indicate that recent methods for part discovery seem to be tailored to datasets with specific characteristics: Dino ViT~\cite{amir_deep_2022_dinovit} thrives with a diverse set of natural images belonging to different supercategories such as PartImageNet but fails to provide the sought after results on the more narrow CelebA, where the parts of interest are restricted to facial landmarks, and the opposite is true for~\cite{huang_interpretable_2020}. Our proposed method, on the other hand, is able to extract semantically consistent parts on all tested datasets without the need for any dataset-specific adjustment, showing its potential for out-of-the-box application to datasets with different characteristics.


\section{Conclusion}

We propose a method for fine-grained visual categorization that uses part representations as an information bottleneck and thus learns to detect semantically consistent parts that are useful for that task. 
Our method requires no additional annotation effort and leverages the fine-grained class labels as the sole supervision signal.
The quantitative and qualitative comparisons against recent part discovery methods shows that our approach improves upon the state-of-the-art in part localization and semantic consistency, with parts specializing in certain categories and consistently overlapping with the same semantic elements of the objects of interest, without sacrificing accuracy on the down-stream classification task. 

There are several directions in which more work is needed to improve this approach. The first relates to the fact that, by applying a mask to a high-level feature map in a deep model, we have no guarantee that only the underlying regions of the image influence the corresponding part feature representation. Information from the background or neighboring parts can leak into the feature representation of a part due to the large receptive field of most modern architectures, limiting the interpretability of the approach.
In addition to this, our results show that PDiscoNet displays a lower level of contour adherence than a method trained with a very large dataset with self-supervision. This, in turn, could affect the interpretability of the part maps and allow background information to substantially affect the part feature representation.

We hope that this approach will contribute towards making models for fine-grained visual categorization more interpretable by facilitating inspection of some aspects of the model's internal reasoning, thus allowing a much richer interaction between the model and its end users. 

\section*{Acknowledgements} This work was supported by the French National Research Agency under the Investments for the Future Program, referred as ANR-16-CONV-0004 (DigitAg), by the ERC (853489 - DEXIM), by the DFG (2064/1 – Project number 390727645), by the Tübingen AI Center (BMBF, FKZ: 01IS18039A), and by the MUR PNRR project FAIR - Future AI Research (PE00000013) funded by the NextGenerationEU.

{\small
\bibliographystyle{ieee_fullname}
\bibliography{egbib}
}

\end{document}